\documentclass[review, sort&compress, 3p]{elsarticle}

\usepackage{hyperref}
\usepackage{makeidx}
\usepackage{color,xcolor}

\usepackage{amsmath,amssymb,amsfonts,amsthm}
\usepackage{mathrsfs}%
\usepackage{unicode-math}
\let\bm\symbf
\usepackage{mathtools}
\usepackage{algorithm,algorithmicx}%
\usepackage{algpseudocode}%
\usepackage{listings}%

\usepackage[mathlines,displaymath]{lineno}
\modulolinenumbers[1]

\usepackage{graphicx}
\usepackage{subfigure}
\usepackage{epsfig}
\usepackage{tikz}
\usetikzlibrary{arrows, snakes, patterns, backgrounds}
\usepackage{ifthen}

\usepackage{tabularx, multirow, arydshln}	
\usepackage{footnote}	
\usepackage{array}

\usepackage{enumerate}
\usepackage{verbatim}

\usepackage{textcomp}%
\usepackage{manyfoot}%
\usepackage{booktabs}%
\usepackage{xeCJK}
\setCJKmonofont{FandolFang-Regular.otf}

\newcommand*{\abs}[1]{\left|#1\right|}

\newcommand{\trans}{\text{T}}

\newcommand{\Phase}[1]{%
	\State \textsc{#1:}%
	\setcounter{stepcount}{0}%
}
\newcommand{\Step}[1]{%
	\stepcounter{stepcount}%
	\State \hspace{2em}\arabic{stepcount}.\ \textbf{#1}%
	\setcounter{substepcount}{0}
}
\newcommand{\Substep}{%
	\stepcounter{substepcount}%
	\State \hspace{4em}\alph{substepcount}.\ %
}
\newcounter{stepcount}
\newcounter{substepcount}

\begin{document}


\begin{frontmatter}
	\title{Adaptive multi-resolution Gaussian processes:
	Scalable exact inference with naturally data-sparse covariance matrices}

	\author{Yanchuang Cao (曹衍闯)}
	\author{Jun Liu (刘军)}
	\author{Tengchao Yu (于腾超)\corref{cor}} \ead{yu\_tengchao@iapcm.ac.cn}
	\author{Heng Yong (勇珩)}

	\cortext[cor]{Corresponding author}
	\address{Institute of Applied Physics and Computational Mathematics, 
	Beijing 100094, China}

	\begin{abstract}
		Gaussian processes constitute a cornerstone of probabilistic machine 
		learning, yet scaling them to large datasets typically forces a trade-off
		between computational efficiency and model fidelity.
		This work bridges this gap by presenting an
		adaptive multi-resolution Gaussian process framework that is
		both scalable and exact.
		Our key innovation is constructing a naturally data-sparse covariance 
		matrix with adaptive multi-resolution basis functions.
		These basis functions are directly anchored to samples, 
		eliminating the need for auxiliary points.
		By shrinking the support domains of multi-resolution basis, the 
		matrix block sizes are limited, guaranteeing sparsity.
		The inverse of the data-sparse covariance matrix is computed 
		exactly and efficiently via the sparse Cholesky inverse algorithm.
		To further improve predictive uncertainties, we construct an augmented 
		basis function.
		Theoretical analysis and numerical experiments demonstrate that our
		model achieves exact inference with $\mathcal{O}(n \log^2 n)$
		training cost and $\mathcal{O}(\log^d n)$ prediction cost, 
		establishing a principled framework for scalable and high-fidelity
		Gaussian process regression.
	\end{abstract}
\end{frontmatter}

\section{Introduction}

Gaussian processes (GPs) provide a powerful approach for probabilistic 
machine learning \citep{ghahramani2015probabilistic, cruz2024survey} due to its
explainability, flexibility, and rigorous Bayesian foundations.
It has been applied in various communities, such as 
geostatistics \citep{liuxu2021multiresolution},
probabilistic numerical simulation \citep{chenyifan2025sparse}, 
Bayesian optimization \citep{shahriari2016review}, 
uncertainty quantification \citep{manfredi2024uncertainty, lijinglai2025gpr4uq}, etc.
Their popularity stems from explainability, flexibility, and rigorous Bayesian
foundations.
However, standard GPs require solving linear systems with $n \times n$ fully 
populated covariance matrices $\bm{K}$ for $n$ training samples, 
resulting in $\mathcal{O}(n^3)$ computational complexity for direct solvers and 
$\mathcal{O}(n_\mathrm{iter} n^2)$ for iterative solvers \citep{rasmussen2006gp}.
This high complexity makes standard GPs prohibitive for large scale problems.

Much effort has been devoted to developing scalable GP models with 
low computational complexity in the past decades.
Most of them are based on approximations for the exact GP model
\citep{liuhaitao2020review,
heaton2019competition, cressie2022basis, dongkun2017determinants}.
Popular methods generally fall into the following broad categories.
\begin{itemize}
	\item \textbf{Grid-based methods:}
		For stationary kernels $k(\bm{x}, \bm{x}') = k(\bm{x}-\bm{x}')$ which is 
		translational invariant, if the training samples lie on Cartesian grids, 
		the covariance matrix $\bm{K}$ would become Toeplitz and can be efficiently
		via fast Fourier transforms (FFT), resulting in $\mathcal{O}(n \log n)$ 
		complexity \citep{dietrich1997fastgp}.
		For arbitrarily distributed samples, it can also be accelerated by
		projecting data between data points and grid points, leading to the well 
		known structured kernel interpolation (SKI) method and its variants 
		\citep{wilson2015kissgp, banhanyuan2024mkissgp, greengard2025efgp, 
		kielstra2025gp}.
		When applied to multi-dimensional problems, however, the Cartesian grids
		consists of $m^d$ points and requires to $(n + d m^d n \log m)$.
		Even with fixed $m$, the computation can becomes prohibitive since 
		it grows exponentially with dimension.
		To alleviate such curse of dimensionality, 
		the structured kernel interpolation for products (SKIP) is proposed
		\citep{gardner2018skip} which approximate the high-dimensional kernel as 
		the product of a set of one-dimensional kernels and brings down the 
		complexity to $\mathcal{O}(dn + dm \log m)$.
	\item \textbf{Mixture of local experts:}
		Generally the target function value at the test point is more relevant 
		with neighboring samples, thus it can be predicted approximately with 
		the local Gaussian process model considering only its neighborhoods.
		This idea leads to local approximate Gaussian processes (laGP)
		\citep{gramacy2015lagp, gramacy2016lagp, fuhg2022lagp}.
		This is equivalent to approximating the covariance matrix with
		a band-diagonal one.
		Another approach is first divide the training points into clusters and
		train separate GPs which are experts at corresponding local regions, 
		then the prediction is given by combining the predictions of these 
		local experts \citep{tuong2009lgp, gaoyinghua2020lgp, rumsey2023leapgp,
		adjetey2026jump}.
		Thus the covariance matrix is approximated by a function of its diagonal
		blocks.
		For multi-dimensional problems, it may suffers from the drawback that 
		the size of neighborhoods or clusters have to increase exponentially 
		with dimension to give good predictions \citep{gramacy2015lagp}.
	\item \textbf{Low rank approximations:}
		Smooth kernels are used in most GP models,
		thus the matrix $\bm{K}$ can be approximated by low rank approximations
		$\bm{K} \approx \bm{S} \bm{D} \bm{S}^\top$, with $\bm{S}$ being an 
		$n \times r$ matrix and $\bm{D}$ being an $r \times r$ matrix.
		Thus the complexity becomes $\mathcal{O}(n r^2)$ and can be greatly reduced 
		when $r \ll n$.
		Various techniques can be used to construct such an approximation.
		For example, the sparse on-line Gaussian processes selects $r$ representative
		samples \citep{csato2002sgp},
		sparse spectrum Gaussian processes (SSGP) approximates the kernel with 
		$r$-order expansions in the frequency domain \citep{gredilla2010ssgp, 
		tan2016ssgp},
		fixed rank kriging (FRK) gets the low rank approximation by using $r$ basis 
		functions \citep{cressie2008frk, mangion2021frk}, 
		and a group of sparse Gaussian processes create pseudo-inputs at $r$ 
		inducing points \citep{snelson2006spgp, titsias2009sgpr, hensman2013gp, 
		ketenci2025cvgp}. 
		The major drawback of these methods is that they tend to capture 
		global trends only and may have difficulty in predicting
		local fluctuations.
	\item \textbf{Hierarchical approximations:}
		Implementing low rank approximations hierarchically is an attracting 
		scheme to get scalable GPs suitable for capturing multi-scale features.
		Generally the kernel is smooth and decays with distance, thus the covariance
		matrix $\bm{K}$ is hierarchically off-diagonal low-rank (HODLR),
		and its approximate factorization requires only $\mathcal{O}(n \log^2 n)$ 
		computations \citep{ambikasaran2016gp}.
		It is efficient with high accuracy for one-dimensional cases, but
		its performance can noticeably degrade even for $d=2$ since the rank 
		grows with dimension \citep{minden2017sgp, kielstra2025gp}.
		Multi-resolution Gaussian processes (MRGP) constructs the hierarchical
		low rank approximation by dividing the region
		recursively and constructing an efficient
		sub-model with carefully chosen auxiliary points on each subregion
		\citep{fox2012multiresolution, taghia2019multiresolution, 
		nychka2015multiresolution, katzfuss2017multiresolution, 
		katzfuss2020multiresolution, huanghuang2019mrgp}.
		Study on its applications to one and two dimensional problem shows that
		it is often appropriate to divide the region $\log n$ times with 
		$\mathcal{O}(1)$ auxiliary points used in each sub-model, which also leads to
		$\mathcal{O}(n \log^2 n)$ complexity \citep{katzfuss2020multiresolution}.
		Theoretically it is equivalent to the HODLR-based method to some extent, 
		thus it would suffer from the same drawback when applied to 
		multi-dimensional problems, for which more auxiliary points
		would be required as $d$ increases.
	\item \textbf{Vecchia approximations:}
		It is pointed out that based some ordering of the samples, 
		both the Cholesky factorization of the covariance matrix $\bm{K}$ and 
		its inverse are approximately sparse.
		The sparsity can be identified by selecting neighboring samples from 
		subsets of training samples, and the matrix entries can be computed 
		via minimizing the Kullback-Leibler divergence between Gaussian 
		distributions with approximate and exact covariance matrices.
		Representative methods includes the nearest neighbor Gaussian processes
		(NNGP) \citep{datta2016nngp, finley2019nngp, finley2022spnngp},
		scaled Vecchia approximation \citep{katzfuss2022svecgp}, 
		and the sparse Cholesky factorization for GPs \citep{schafer2021compression, 
		schafer2021sparse, guinness2018permutation, chenyifan2025sparse}.
		Katzfuss presents a general framework for Vecchia approximations, and 
		demonstrates that the multi-resolution approximation (MRA) 
		can also be viewed as a special case of the Vecchia approximation
		\cite{katzfuss2021vecchia}.
		Generally Vecchia approximations gives predictions within log-linear
		computational complexity, but the cost may grows with dimension 
		to maintain its accuracy.
		For example, the computational cost of sparse Cholesky factorization
		for GPs grows at the speed of $\mathcal{O}(n \log^d (n/\varepsilon))$
		with $\varepsilon$ being the predetermined accuracy 
		\cite{chenyifan2025sparse}.
\end{itemize}
Some scalable approximate GPs may cannot be grouped into the above categories,
see \cite{liuhaitao2020review, heaton2019competition} and \cite{rumsey2025emulators} 
for recent review and comparison of state-of-the-art models.

Although approximate GPs can substantially reduce the computational cost and 
have gained great success in various fields, 
the approximation error introduces additional computational uncertainty
in prediction \citep{wenger2022uncertainty}.
This necessitates a fundamental trade-off between computational efficiency
and model fidelity.
For high-fidelity modeling, particularly when capturing highly non-linear
functions, the required number of inducing points or neighboring points 
has to approach the full dataset size.
Consequently, the computational advantages of these scalable GP approximations
are effectively negated \citep{wangke2019exactgp, rumsey2023leapgp}.
Furthermore, most approximate GPs rely on stationary and smoothly decaying
kernel functions, which may inherently constrain their ability to achieve
high-fidelity for highly non-linear functions.

Scalable exact GPs, on the contrary, try to construct sparse GP models with
carefully designed kernels rather than construct sparse approximations
for the standard GP model.
Notice here ``exact'' means precisely solving $\bm{K}_y^{-1}$ in the constructed
GP model to give exact inference, rather than exact inference for the standard 
GP model with common kernels.
Therefore, they require no trade-off between computational efficiency and
model fidelity, and no computational uncertainty is induced.
A straightforward idea is to replace the globally supported kernel with a 
compactly supported one by tapering, resulting in a band diagonal covariance 
matrix \citep{furrer2006tapering, shaby2012tapering}.
The compactly supported kernel may be learned from the data rather than
designed in advance \citep{noack2023nonstationary}, resulting in a non-stationary
and flexible covariance function and theoretically better predictions.

Another category of scalable exact models consider GPs in the weight-space view,
and construct models with compactly supported basis.
Typical examples include the relevance vector machine (RVM)
\citep{tipping2001rvm, rasmussen2005healing}
and the actually sparse GP \citep{cunningham2023sparse}, in which
the selected basis are supported only on local regions.
According to the relationship between basis and the covariance kernel,
the resulting covariance matrices are also sparse and can be computed
efficiently.

Theoretically, scalable exact GPs can be applied to high-fidelity modeling
with big data since exact inference are computed.
However, the quality of predictions depends on the kernel
designed in these methods, since the kernel actually reflects the prior
knowledge of the target function.
Kernels in most scalable exact GPs are stationary and defined at a single, 
fixed scale,
which fundamentally limits their flexibility in capturing multi-scale features.
The sparsity-discovering kernel in \cite{noack2023nonstationary} learns
non-stationary kernels from data that can represent relevance of points 
at distance, but its
computational cost is remarkably much higher than approximate GPs.
These may be the reasons why they are much less popular than 
approximate GPs in the community of scalable GPs.

To bridge the key gap between the approximation-based scalability and 
high-fidelity modeling for highly non-linear functions, 
we propose a scalable exact GP framework with adaptive multi-resolution basis
in this paper.
The core of our adaptive multi-resolution Gaussian process (AMRGP) model 
is to construct a naturally data-sparse covariance
matrix with adaptive multi-resolution basis.
That is, although the covariance matrix is fully populated, but 
it can be represented with few data.
The basis functions are directly anchored to samples, 
eliminating the need for auxiliary points.
By shrinking the support domains of multi-resolution basis, the matrix block sizes 
are limited and the matrix sparsity is guaranteed.
The inverse of the data-sparse covariance matrix is then computed 
exactly and efficiently via the sparse Cholesky inverse algorithm.
Moreover, we construct an augmented basis function to 
provide reasonable predictive uncertainties.
Exact inference is achieved
with $\mathcal{O}(n \log^2 n)$ training cost and 
$\mathcal{O}(\log^d n)$ prediction cost.
Thus our AMRGP framework offers a fundamentally more principled 
path to scalability, 
where efficiency arises from the model's exact sparse representation
rather than from approximations of a dense model.

The rest of the paper is organized as follows.
Section \ref{sec_framework} describes the main framework of our AMRGP.
Section \ref{sec_basis} gives the construction of the 
adaptive multi-resolution basis anchored to samples, based on which the sparse
Cholesky inverse is implemented.
Then we analyse its computational complexity in Section \ref{sec_complexity}.
Finally, Section \ref{sec_numexp} studies the performance of our model
with numerical examples.

\section{Scalable exact Gaussian process framework with multi-resolution basis}
\label{sec_framework}

The basic idea of our method begins with interpreting GPs in the weight-space
view, which would be briefly introduced in Section \ref{subsec_gp}.
Then we demonstrate in Section \ref{subsec_mrgp} that with multi-resolution
basis, the covariance matrix is naturally data-sparse,
and the GPs can be computed efficiently with sparse Cholesky inverse.
Finally, in Section \ref{subsec_aug} we give the augmented process to 
provide reasonable predictive uncertainties.

\subsection{Gaussian processes in the weight-space view}
\label{subsec_gp}

This subsection gives a brief overview of Gaussian processes in the 
weight-space view. Readers may refer to \citep{rasmussen2006gp} for 
more details.

For the prediction of a function $f(\bm{x})$, generally it may be approximated by
\begin{equation} \label{eq_wgtexp}
	f(\bm{x}) = \bm{\phi}^\top(\bm{x}) \bm{w}
\end{equation}
with basis functions 
$\bm{\phi}(\bm{x}) = \{ \phi_i(\bm{x}) \}$ and their weights $\bm{w} = \{w_i\}$.
GP regression aims to provide a probabilistic prediction of $f(\bm{x})$ via
Bayesian inference with $n$ samples
$(\bm{X}, \bm{y}) = \{ (\bm{x}_i, y_i) \}$,
where $y_i = f(\bm{x}_i) + \varepsilon_i$ are observations of the target function
with Gaussian noise $\bm{\varepsilon} \sim \mathcal{N}(0, \sigma_n^2\bm{I})$.

The Bayesian inference requires the prior knowledge of the weights' distribution.
Generally it can be assumed that $\bm{w} \sim \mathcal{N}(\bm{0}, \bm{I})$.
Then the covariance of $f(\bm{x})$ and $f(\bm{x}')$ is
\begin{equation} \label{eq_cov}
	k(\bm{x}, \bm{x}') = \bm{\phi}^\top(\bm{x}) \bm{\phi}(\bm{x}').
\end{equation}
Thus the joint distribution of the observed $\bm{y}$ and $y_\ast$ 
at the test point $\bm{x}_\ast$ under the prior is
\begin{equation}
	\left[ \begin{matrix}
		\bm{y} \\
		y_\ast
	\end{matrix} \right] \sim 
	\mathcal{N}\left(\bm{0}, 
	\left[ \begin{matrix}
		k(\bm{X}, \bm{X}) + \sigma_n^2 \bm{I} & k(\bm{X}, \bm{x}_\ast) \\
		k(\bm{x}_\ast, \bm{X}) & k(\bm{x}_\ast, \bm{x}_\ast)
	\end{matrix} \right] \right) =  \mathcal{N} \left( \bm{0}, 
	\left[ \begin{matrix}
		\bm{K}_y & \bm{k}_\ast \\
		\bm{k}_\ast^\top & k_{\ast\ast}
	\end{matrix} \right] \right)
\end{equation}
where $\bm{K}_y = \bm{K} + \sigma_n^2 \bm{I}$ and
\begin{equation}
	\bm{K} = \bm{\Phi}^\top \bm{\Phi}, \quad 
	\bm{k}_\ast = \bm{\Phi}^\top \bm{\phi}_\ast
\end{equation}
with $\bm{\Phi} = \{ \phi_i(\bm{x}_j) \}$ and $\bm{\phi}_\ast = \{ \phi_i (\bm{x}_\ast) \}$.
The probabilistic prediction of $f(\bm{x})$ can be derived by computing the
conditional distribution of $f_\ast = f(\bm{x}_\ast)$
\begin{equation}
	f_\ast | \bm{x}_\ast, \bm{X}, \bm{y} \sim \mathcal{N}(\mu_\ast, \sigma_\ast^2)
\end{equation}
with
\begin{equation} \label{eq_predk}
	\begin{split}
		\mu_\ast =& \bm{k}_\ast^\top \bm{K}_y^{-1} \bm{y}, \\
		\sigma_\ast^2 =& k_{\ast\ast} - \bm{k}_\ast^\top \bm{K}_y^{-1}
		\bm{k}_\ast.
	\end{split}
\end{equation}
The prediction $\mu_\ast$ may also be written as
\begin{equation} \label{eq_expectation}
	\mu_\ast = \bm{\phi}_\ast^\trans \bm{\omega}
\end{equation}
with 
\begin{equation}
	\bm{\omega} = \bm{\Phi}\bm{K}_y^{-1}\bm{y}
\end{equation}
being the expectation of the weights.
The log marginal likelihood (LML) of the prediction can be evaluated by the 
following formulae when necessary.
\begin{equation} \label{eq_lml}
	\log p(\bm{y} | \bm{X}) = 
	-\frac{1}{2} \bm{y}^\top \bm{K}_y^{-1} \bm{y} 
	- \frac{1}{2} \log \left| \bm{K}_y \right|
	- \frac{n}{2} \log 2\pi.
\end{equation}
Therefore, the key of scalable GPs is to construct an efficient algorithm
for $\bm{K}_y^{-1} \bm{y}$ computation.

\subsection{Multi-resolution Gaussian processes with naturally data-sparse 
covariance matrices}
\label{subsec_mrgp}

Generally, scalable exact GPs tries to get a sparse covariance matrix 
$\bm{K}$. This may be achieved by designing a locally supported kernel
$k(\bm{x}, \bm{x}')$ or defining a GP model with locally supported basis.
Although these models can be efficient, but theoretically they may have 
difficulty in capturing the global trend since the relevance of points
at distance is ignored.
To overcome this drawback, multi-resolution basis is used in our 
framework which can capture hierarchical features at multi-scales.

When multi-resolution basis is used, clearly $\bm{\Phi}$ is sparse,
but covariance matrix $\bm{K} = \bm{\Phi}^\trans \bm{\Phi}$ is still 
fully populated. 
The reason is that, generally the support domain of the 
coarsest scale basis functions covers the entire region, thus the entries
\begin{equation}
	k(\bm{x}_i, \bm{x}_j) = \sum_{k=1}^n \phi_k(\bm{x}_i) \phi_k(\bm{x}_j)
\end{equation}
are generally non-zero, since both $\phi_k(\bm{x}_i)$ and $\phi_k(\bm{x}_j)$
are non-zero when $\phi_k(\bm{x})$ is at the coarsest scale.
Therefore, $\bm{K}_y = \bm{\Phi}^\trans \bm{\Phi} + \sigma_n^2 \bm{I}$ 
is data-sparse but not explicitly sparse.
The direct computation of $\bm{K}_y^{-1}$ is 
still computational intensive, and we must find an efficient method for
the computation.

Notice $\bm{K}_y^{-1}$ can be computed by the 
Sherman-Morrison-Woodbury formula,
\begin{equation}
	\bm{K}_y^{-1} = (\bm{K} + \sigma_n^2 \bm{I})^{-1} 
	= \sigma_n^{-2} (\bm{I} - \bm{\Phi}^\trans \bm{G}^{-1} \bm{\Phi})
\end{equation}
where
\begin{equation}
	\bm{G} = \bm{\Phi} \bm{\Phi}^\trans + \sigma_n^2 \bm{I}.
\end{equation}
Most functions in the set of multi-resolution basis are locally supported, thus
$\bm{G}$ is actually sparse since 
\begin{equation}
	G_{ij} = \sum_{k=1}^n \phi_i(\bm{x}_k) \phi_j(\bm{x}_k) + \sigma_n^2 \delta_{ij}
\end{equation}
is zero when the support domains of $\phi_i$ and $\phi_j$ have no overlapping
regions.
Based on some ordering of the basis, the Cholesky factorizations of 
$\bm{G} = \bm{L}\bm{L}^\trans$ and its inverse
$\bm{G}^{-1} = (\bm{L}^{-1})^\trans \bm{L}^{-1}$ are also sparse and can be
computed efficiently \citep{brandhorst2011sparse}.
In the next section we will see that the product $\bm{L}_\phi = \bm{L}^{-1} \bm{\Phi}$
may also be sparse and can be computed efficiently. Thus
\begin{equation}
	\bm{K}_y^{-1} = \sigma_n^{-2} (\bm{I} - \bm{L}_\phi^\trans \bm{L}_\phi)
\end{equation}
and the computational cost of the prediction can be reduced.

\subsection{Augmented Gaussian processes with compactly supported basis}
\label{subsec_aug}

In this subsection, we will point out that GP models constructed with
compactly supported basis tends to severely underestimate the predictive 
uncertainty, and this drawback can be alleviated by augmentation.

In the weight-space view, GP model gives a probabilistic prediction
for $f(\bm{x})$ by computing the weights $\bm{w}$ in 
$f(\bm{x}) = \bm{\phi}^\top (\bm{x}) \bm{w}$.
When all the basis functions are linearly independent and the amount
is exactly identical to the observations, 
the system matrix $\bm{\Phi}$ is invertible.
Therefore, if there is no noise in the observations, 
the weights $\bm{w}$ would be deterministic and the uncertainties would vanish.
More generally, the prediction uncertainty can be computed by
\begin{equation} \label{eq_uncertainty}
	\sigma_\ast^2 = \sigma_n^2 \bm{\phi}_\ast^\top \bm{G}^{-1} \bm{\phi}_\ast.
\end{equation}
Thus when $\bm{G} = \bm{\Phi} \bm{\Phi}^\top + \sigma_n^2 \bm{I}$ is 
non-singular, the uncertainty of the prediction only comes from
the noise in the observation.
In other words, only aleatory uncertainty is considered.
The epistemic uncertainty, which arises from of lack of knowledge of
the target function and incomplete data, is ignored.
Thus the given predictive uncertainty in unobserved regions is severely
underestimated.

Such a contradiction comes from the prior assumption 
$f(\bm{x}) = \bm{\phi}^\top (\bm{x}) \bm{w}$ with $n$ basis functions,
which indicates that we already have enough knowledge and $n$
observing data is sufficient to reconstruct $f(\bm{x})$.
But in fact, the target function may be in arbitrarily form,
and numerous basis functions have to be used.
Thus an under-determined linear system with infinite solutions should be established.
The distribution of the weights over its solution space
gives the probabilistic prediction for the target function.
With such over-complete basis, $\bm{G}$ in the uncertainty estimation 
\eqref{eq_uncertainty} approaches singular when $\sigma_n^2$ vanishes, thus 
the prediction uncertainty $\sigma_\ast^2$ can still have finite values even 
there is nearly no noise in the observations.
Therefore, more that $n$ basis functions should be used to ensure the prediction 
uncertainty does not go to zero.
Adding additional basis functions are called augmentation in this work.

For simplicity and efficiency, we use only one augmented basis function
$\phi^\ast(\bm{x})$ in this work.
Moreover, $\phi^\ast(\bm{x})$ is carefully designed to make it zero
at all training points, thus the covariance matrix $\bm{K}$ and the 
prediction mean $\mu_\ast$ remains exactly the same with the unaugmented GP.
The augmented matrix $\bm{G}^\ast$ is
\begin{equation}
	\bm{G}^\ast  = \left[ \begin{matrix}
		\bm{\Phi} \\
		\bm{0}
	\end{matrix} \right]
	\left[ \begin{matrix}
		\bm{\Phi} \\
		\bm{0}
	\end{matrix} \right]^\top
	+ \sigma_n^2 \bm{I}
	= \left[ \begin{matrix}
		\bm{\Phi}\bm{\Phi}^\top + \sigma_n^2 \bm{I} & \\
		& \sigma_n^2
	\end{matrix} \right]
	= \left[ \begin{matrix}
		\bm{G} & \\
		& \sigma_n^2
	\end{matrix} \right].
\end{equation}
Thus the uncertainty becomes
\begin{equation} \label{eq_uncer_aug}
	\sigma_\ast^2 = \sigma_n^2 \bm{\phi}_\ast^\top \bm{G}^{-1} \bm{\phi}_\ast
	+ \left[ \phi^\ast(\bm{x}_\ast) \right]^2.
\end{equation}
Thus it consists of two terms, emulating the aleatory uncertainty arising from
data noise and epistemic uncertainty caused by lack of knowledge and incomplete
data, respectively.
Theoretically the epistemic uncertainty $[\phi^\ast(\bm{x}_\ast)]^2$ should be 
independent with the observation noise, and
grows with the distance to the training data points.
The augmented basis function may be designed as
\begin{equation} \label{eq_augbss}
	\phi^\ast(\bm{x}) = \prod_{i=1}^n 
	\left[1 - W_i(\bm{x}) \right]^\alpha,
\end{equation}
with $W$ being the radial basis function which equals unity at $\bm{x}_i$ and 
approaches to zero at infinity.
$\phi^\ast(\bm{x})$ is zero at all training points and grows with the 
distance to the training data points.
$\alpha$ is used to tune the increasing speed of the uncertainty with
distance from training data points.

\section{Adaptive multi-resolution basis and sparse matrices}
\label{sec_basis}

Our multi-resolution GP regression is adaptive in the sense that they are 
directly anchored to samples rather than Cartesian grid points as 
\cite{nychka2015multiresolution} or user-defined knots as 
\cite{katzfuss2017multiresolution}.
In order to limit the matrix block sizes and guarantee the sparsity,
support domains of multi-resolution basis are shrunk adaptively.
All the basis functions are smooth without truncation at subregion boundaries, thus
the unphysical discontinuities in MRA \cite{katzfuss2017multiresolution} is 
avoided.
Then all the matrices in the computations are data-sparse when the basis functions
are collected in the nested dissection order.
In this section, we first construct the adaptive 
multi-resolution basis, with their support domains shrunk to ensure the sparsity, 
then discuss the nested dissection reordering, and gives the
sparsity pattern for all matrices in our multi-resolution GP regression.

\subsection{Definition of the adaptive multi-resolution basis}

The choice of basis functions has great influence on the efficiency
and model quality.
Bisquare and Wendland basis functions are two popular choices
\citep{cressie2022basis}, among them we select the Wendland function
for its higher-order continuity.
The basis functions in our model are defined as scaled Wendland functions
anchored to samples
\begin{equation}
	\phi(\bar{r}) = s^{1/2} W(\bar{r})
\end{equation}
with
\begin{equation} \label{eq_wendland}
	W(\bar{r}) = \begin{cases}
		(1 - \bar{r})^6 (35 \bar{r}^2 + 18 \bar{r} + 3)/3, 
		&\quad \bar{r} < 1, \\
		0, &\quad \bar{r} \ge 1,
	\end{cases}
\end{equation}
being the Wendland function and
\begin{equation}
	\bar{r} = \frac{1}{s} \text{dist}(\bm{x}, \bm{x}'), \quad
	\text{dist}(\bm{x}, \bm{x}') = \sqrt{\sum_{i=1}^d (x_i - x'_i)^2}.
\end{equation}
Where, $\bm{x}'$ denotes the sample point at which $\phi$ is anchored, and 
$s$ is the size of the support domain.

To construct multi-resolution basis with support sizes
ranging from the coarsest to the finest scale,
they may be defined by the maximum-minimum distance
ordering (maximin ordering) which is widely used in Vecchia's approximations
\citep{schafer2021sparse, chenyifan2025sparse}.
Let $\bm{I}=\{1, 2, \cdots, n\}$ denote the index set of all $n$ training samples.
First pick one point $\bm{x}_{i_1}$ as the first ordered sample.
Then find the point furthest away from the ordered points as the next one,
\begin{equation}
	i_{q+1} = \arg\max_{i\in I\backslash\{i_1, i_2, \cdots, i_q\}}
	\text{dist}\left( \bm{x}_i, \{ \bm{x}_{i_1}, \bm{x}_{i_2}, \cdots, 
	\bm{x}_{i_q} \} \right).
\end{equation}
The support size of the anchored basis is defined as
\begin{equation}
	s_{i_{q+1}} = \rho \text{dist}\left( \bm{x}_{i_{q+1}}, \{ \bm{x}_{i_1}, 
	\bm{x}_{i_2}, \cdots, \bm{x}_{i_q} \} \right).
\end{equation}
with $\rho$ being the hyper-parameter tuning the support size.
The support size of the first ordered sample may be defined as 
$s_{i_1} = s_{i_2}$.

\begin{figure}[ht]
	\centering
	\subfigure[Adaptive multi-resolution bases for 10 random points.] {
		\includegraphics[width=0.45\textwidth]{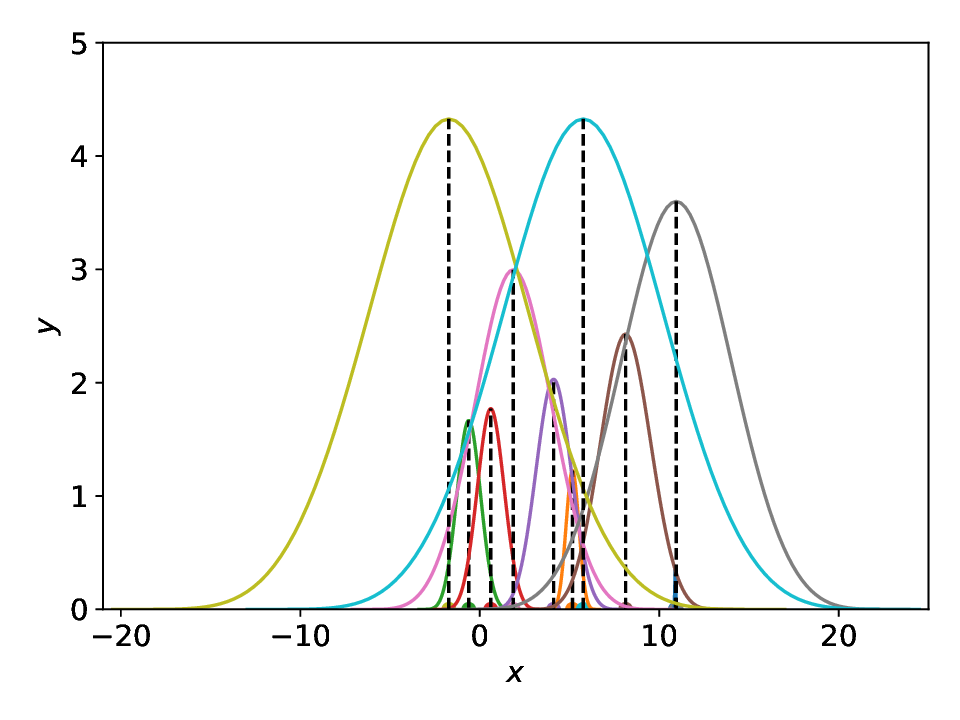}
	}
	\subfigure[Support sizes for 100 random samples.] {
		\includegraphics[width=0.45\textwidth]{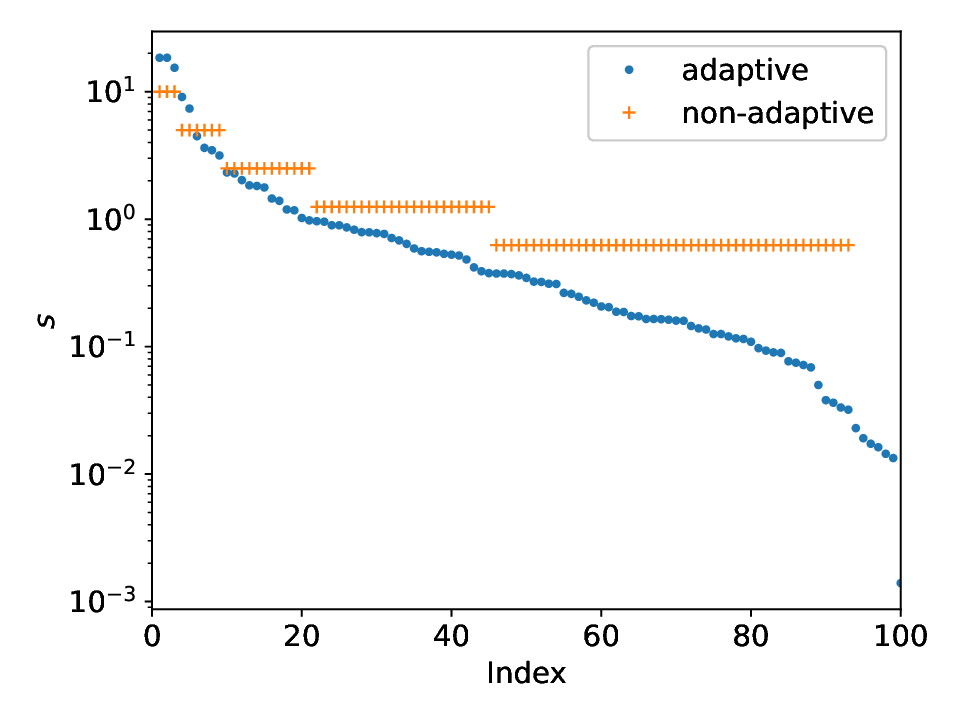}
	}
	\caption{Examples of our adaptive multi-resolution basis for samples with
	$\rho = 2.5$ in $[0, 10]$.}
	\label{fig_basis}
\end{figure}

An example of our adaptive multi-resolution basis is illustrated in 
Fig.~\ref{fig_basis}, and the support sizes are compared with typical 
non-adaptive multi-resolution basis in MRA \cite{katzfuss2017multiresolution}.
The non-adaptive multi-resolution basis is constructed by partitioning the region
up to the fourth level, and three basis functions are defined for each partition.
It is shown that the support sizes of non-adaptive basis vary in a stepwise 
manner, while those of our basis vary approximately smoothly from the coarsest 
to the finest scale.
This position-dependent scaling naturally yields a non-stationary kernel
$k(\bm{x}_i, \bm{x}_j) = \sum_{k=1}^n \phi_k(\bm{x}_i) \phi_k(\bm{x}_j)$, 
since it depends on the absolute positions $\bm{x}_i$ and 
$\bm{x}_j$ rather than on their differences $\bm{x}_i - \bm{x}_j$, 
whereas scalable GPs generally rely on stationary kernels.
Moreover, our basis are straightforwardly anchored to these irregularly distributed 
samples.
Therefore, our adaptive multi-resolution basis tends to be better adapted to 
capturing features at any scale and arbitrarily distributed data points.

It is worth noting that the overlap of the support regions of the basis functions 
may lead to high collinearity among columns of $\bm{\Phi}$, resulting in a large
condition number of the covariance matrix. This ill-conditioning can cause
instability in the regression. In this work, we mitigate this issue 
by simply setting a moderate noise level, which provides sufficient regularization
to stabilize the computation while preserving the model fidelity.
More advanced techniques could be explored in the future.

\subsection{Basis construction and support domain shrinkage}

Constructing basis via maximin reordering straightforwardly 
for all the basis functions requires $\mathcal{O}(n^2)$
computations. An efficient algorithm for the reordering is proposed
in \cite{schafer2021sparse} with $\mathcal{O}(n \log^2 n)$ computational
complexity, but it is quite complicated.
In our AMRGP, the maximin reordering is carried out efficiently on a tree structure.

The adaptive binary tree structure in AMRGP is defined by 
dividing each cluster of samples recursively into two sub-clusters.
In this work, this is carried out via Lloyd's algorithm.
The nested clustering actually gives a hierarchical partition of the region,
with each subregion being a Voronoi polyhedron with no more than $\mathcal{O}(\log n)$
boundary patches.

The basis construction starts from the finest level.
First we define basis functions on samples inside each leaf node straightforwardly 
via the maximin ordering, and further categorize them into
\emph{local} and \emph{non-local} ones.
Here a basis function and the corresponding sample is called \emph{local} 
when its support domain is enclosed
by the node's subregion, otherwise it is called \emph{non-local}.
Whether a basis function is local can be checked by comparing its support size
and the distance between the data point and each boundary patch of the Voronoi
polyhedron, requiring no more than $\mathcal{O}(\log n)$ computations.
Then non-local samples on child nodes are collected to construct local basis
functions on the parent node in the same manner.
This process continues until it reaches the root of the binary tree, where all 
basis functions are local.
For a balanced tree, the construction of multi-resolution basis can be depicted as
\begin{eqnarray*}
	\begin{matrix}
		\bm{X} & \rightarrow & \bm{\psi}_L & \rightarrow & \bm{\psi}_{L-1}
		& \rightarrow & \cdots & \rightarrow & \bm{\psi}_1 & ~ & ~ \\
		~ & \searrow & ~ & \searrow & ~ & \searrow & ~ & \searrow & ~ & \searrow & ~ \\
		~ & ~ & \bm{\phi}_L & ~ & \bm{\phi}_{L-1} & ~ & \cdots & ~ &
		\bm{\phi}_1 & ~ & \bm{\phi}_0
	\end{matrix}
\end{eqnarray*}
Where, $\bm{\psi}_l$ and $\bm{\phi}_l$ denote the \emph{non-local} and \emph{local} 
basis on the $l$-th level, respectively.
All the local basis functions on the binary tree collectively form
the adaptive multi-resolution basis, i.e.,
\begin{equation}
	\bm{\phi} = \{ \bm{\phi}_L, \bm{\phi}_{L-1}, \cdots, \bm{\phi}_0 \}.
\end{equation}

With such adaptive multi-resolution basis, clearly $\bm{\Phi}$ is sparse with
each non-zero matrix block corresponding to interactions of two nodes.
The matrix block size $r$ is identical to the amount of local basis functions
on the corresponding node.
In order to limit the non-zeros and guarantee the sparsity of $\bm{\Phi}$, 
the amount of local basis functions on each node $r$ should be bounded.
However, generally $r$ increases with dimension in the above scheme.
This is because basis functions anchored to samples close to the subregion 
boundary are generally non-local, thus $r$ tends to increase exponentially at 
the speed of $\mathcal{O}(\varrho^{d-1})$ with $\varrho$ being the density of samples
in the space.

To guarantee the sparsity and computational efficiency of matrices, we 
artificially limit the amount of local basis functions on each node with 
$r \le m$. 
Notice that for each node $\mathscr{N}$, the local basis functions are 
constructed from samples when $\mathscr{N}$ is a leaf, or 
from collected non-local samples on its two child nodes when $\mathscr{N}$ is 
not a leaf. Thus $r$ is bounded by $m$ when
\begin{enumerate}
	\item the binary tree structure is constructed by dividing recursively 
		until there are no more than $m$ samples on each leaf node, and
	\item the amount of non-local bases is limited by $m/2$ on each node.
\end{enumerate}

Hence, the key to limiting the amount of \emph{local} basis functions 
is to limit the \emph{non-local} basis functions on each node.
This is done by shrinking the support domain of redundant non-local basis functions.
For each non-local basis functions, suppose it can be transformed into local
by shrinking its support domain
Consider a node with no more than $m$ samples from which the local and non-local 
basis functions to be constructed.
First calculate the support sizes of anchored samples and identify
\emph{naturally local} basis functions with support domains enclosed by the
subregion.
If $q > m/2$ non-local basis functions remain, select $q-m/2$ from them and 
transform them into local basis functions by shrinking their support domains.
In this work, we prioritize transforming basis functions that require minimal 
shrinkage. 
First we compute shrinkage factors $c$ for each non-local basis function
thus with the scaled support size $c_i s_i$, they can be transformed to local.
Then we choose those basis functions with shrinkage
factors closest to 1.0 and transform then into local.

\subsection{Nested dissection ordering and sparse pattern of matrices}

With the adaptive multi-resolution basis constructed above, both
$\bm{\Phi} = [\phi_i(\bm{x}_j)]$ and $\bm{G} = \bm{\Phi}\bm{\Phi}^\trans
+ \sigma_n^2 \bm{I}$ are sparse. Notice in our AMRGP, $\bm{G}^{-1}$ has
to be computed. In this work, it is computed efficiently with the sparse
Cholesky inverse algorithm \citep{brandhorst2011sparse}, which suggests
that if the basis are in the nested dissection order, both the Cholesky
factorization and its inverse are sparse.

\begin{figure} [ht]
	\centering
	\begin{tikzpicture}[>=latex, level distance=30pt]
		\tikzstyle{every node} = [ball color=red!30, circle, text=blue]
		\tikzstyle{level 1} = [sibling distance = 160pt]
		\tikzstyle{level 2} = [sibling distance =  80pt]
		\tikzstyle{level 3} = [sibling distance =  40pt]
		\tikzstyle{level 4} = [sibling distance =  20pt]
		\node {1}
			child { node{2}
				child { node{4} }
				child { node{5} }
			}
			child { node{3}
				child { node{6}
					child { node{8} }
					child { node{9} }
				}
				child { node{7} }
			};
	\end{tikzpicture}
	\caption{For this tree structure, the nodes are collected in nested
	dissection order as $\{4, 5, 2, 8, 9, 6, 7, 3, 1\}$.}
	\label{fig_nd}
\end{figure}
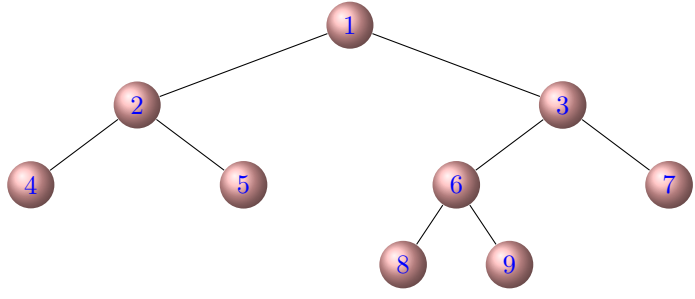

Since all the basis are defined on nodes, thus the reordering can be
carried out by collecting local basis on nodes in the nested dissection ordering.
First define a empty node list, to which all the nodes will be pushed back
in the nested dissection order.
Check whether the root can be pushed
back into the list. For each node under consideration, it is pushed back 
into the list if it is a leaf, or all of its descents are already in the list
if it is a non-leaf node, otherwise check whether its children can be pushed back.
This is done recursively until the root is pushed back into the list.
An example of the reordered nodes for a binary tree is provided
in Fig.~\ref{fig_nd}.

Now let's study the sparse patterns of matrices in the GP model.
First we define some notations for the convenience of discussion.
Consider a node $\mathscr{N}_i$ in the binary tree.
\begin{itemize}
	\item $\mathcal{A}(\mathscr{N}_i)$: the set of all ancestors of node 
		$\mathscr{N}_i$. For example, the set of all ancestors of Node 8
		in Fig.~\ref{fig_nd} is $\{6, 3, 1\}$.
	\item $\mathcal{D}(\mathscr{N}_i)$: the set of direct descendants of node
		$\mathscr{N}_i$. For example, the set of direct descendants of Node 3
		in Fig.~\ref{fig_nd} is $\{6, 7, 8, 9\}$.
	\item $\mathcal{R}(\mathscr{N}_i) = \mathcal{A}(\mathscr{N}_i) \bigcup
		\mathcal{D}(\mathscr{N}_i) \bigcup \{ \mathscr{N}_i \}$: the set of
		direct relatives of $\mathscr{N}_i$. For example, the set of direct
		relatives of Node 6 is $\{1, 3, 6, 8, 9\}$.
\end{itemize}
Clearly, we have
\begin{itemize}
	\item if $\mathscr{N}_i \in \mathcal{A}(\mathscr{N}_j)$, then 
		$\mathscr{N}_j \in \mathcal{D}(\mathscr{N}_i)$, and vice versa;
	\item if $\mathscr{N}_i \in \mathcal{R}(\mathscr{N}_j)$, then 
		$\mathscr{N}_j \in \mathcal{R}(\mathscr{N}_i)$;
	\item if $\mathscr{N}_i \in \mathcal{A}(\mathscr{N}_k)$ and 
		$\mathscr{N}_j \in \mathcal{R}(\mathscr{N}_k)$, then
		$\mathscr{N}_i \in \mathcal{R}(\mathscr{N}_j)$.
\end{itemize}

Now let's discuss the block-wise sparsity of $\bm{\Phi}$.
For each node $\mathscr{N}_j$, clearly the local basis functions may be non-zero on 
local sample points in its descendants since they also lie inside $\mathscr{N}_j$.
It is worth noting that they may also take non-zero values on local sample 
points in its ancestors, since part of them may be the non-local sample points
in $\mathscr{N}_j$.
Therefore, $\bm{\Phi}$ consists of non-zero blocks for $(\mathscr{N}_i, 
\mathscr{N}_j)$ with $\mathscr{N}_i \in \mathcal{R}(\mathscr{N}_j)$.

For the matrix $\bm{G} = \bm{\Phi} \bm{\Phi}^\trans + \sigma_n^2 \bm{I}$, a
block $\bm{G}(\mathscr{N}_i, \mathscr{N}_j)$ corresponding with 
$(\mathscr{N}_i, \mathscr{N}_j)$ is non-zero only when the local samples on
$\mathscr{N}_i$ and $\mathscr{N}_j$ have overlapping support domains.
Since generally the support domains of local samples on a node is enclosed by
the corresponding subregion, this only happens when $\mathscr{N}_i$
and $\mathscr{N}_j$ have overlapping subregions.
Thus $\bm{G}(\mathscr{N}_i, \mathscr{N}_j)$ is non-zero only when 
$\mathscr{N}_i \in \mathcal{R}(\mathscr{N}_j)$.
That is, the sparse pattern of $\bm{G}$ is the same with that of $\bm{\Phi}$.

\begin{figure}[ht]
	\centering
	\subfigure[Sparse pattern of $\bm{\Phi}, \bm{G}$, and $\bm{L}_\phi$.] {
		\includegraphics[width=0.45\textwidth]{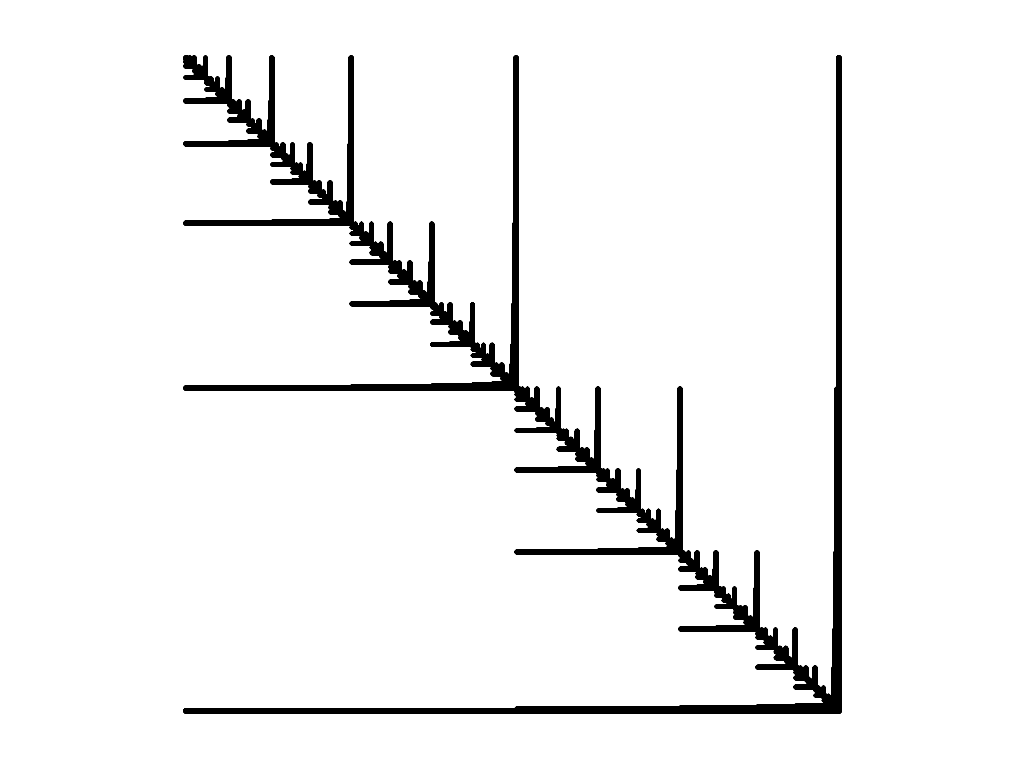}
		\label{fig_ptnG}
	}
	\subfigure[Sparse pattern of $\bm{L}$, and $\bm{L}^{-1}$.] {
		\includegraphics[width=0.45\textwidth]{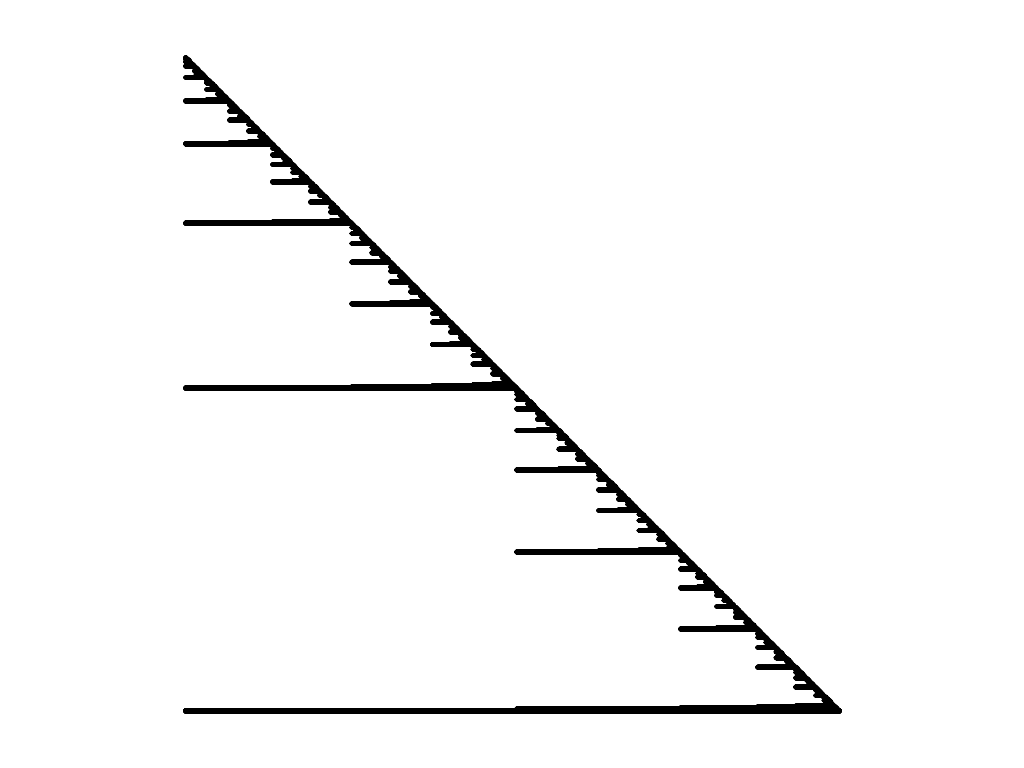}
		\label{fig_ptnL}
	}
	\caption{Sparse pattern of matrices in an AMRGP model with 10,000 samples.}
	\label{fig_pattern}
\end{figure}

The sparse pattern of $\bm{\Phi}$ and $\bm{G}$ is illustrated in Fig.~\ref{fig_ptnG}.
With such a sparse pattern, both $\bm{L}$ in its Cholesky factorization and 
$\bm{L}^{-1}$ are sparse.
They can be computed efficiently with elimination trees 
\citep{brandhorst2011sparse, liuwh1986cholesky}.
That is, for each block corresponding to $(\mathscr{N}_i, \mathscr{N}_j)$
in $\bm{L}$ or $\bm{L}^{-1}$, it is nonzero only when 
$\mathscr{N}_i \in \mathcal{A}(\mathscr{N}_j)$ or $\mathscr{N}_i = \mathscr{N}_j$.
Thus the sparse pattern of $\bm{L}$ and $\bm{L}^{-1}$ is the same with that
of $\bm{G}$, except it consists of only diagonal and sub-diagonal blocks.
The sparse pattern of $\bm{L}$ and $\bm{L}^{-1}$ are depicted in 
Fig.~\ref{fig_ptnL}.

For the matrix $\bm{L}_\phi = \bm{L}^{-1} \bm{\Phi}$, the block corresponding
to $(\mathscr{N}_i, \mathscr{N}_j)$ can be computed by
\begin{equation}
	\bm{L}_\phi(\mathscr{N}_i, \mathscr{N}_j) = \sum_k 
	\bm{L}^{-1}(\mathscr{N}_i, \mathscr{N}_k)
	\bm{\Phi}(\mathscr{N}_k, \mathscr{N}_j).
\end{equation}
Notice $\bm{L}^{-1}(\mathscr{N}_i, \mathscr{N}_k) \neq \bm{0}$ only when
$\mathscr{N}_i \in \mathcal{A}(\mathscr{N}_k) \bigcup \{\mathscr{N}_k\}$,
and $\bm{\Phi}(\mathscr{N}_k, \mathscr{N}_j) \neq \bm{0}$ only when
$\mathscr{N}_j \in \mathcal{R}(\mathscr{N}_k)$.
Therefore, the block $\bm{L}_\phi(\mathscr{N}_i, \mathscr{N}_j) \neq \bm{0}$
only when $\mathscr{N}_i \in \mathcal{R}(\mathscr{N}_j)$.
That is, $\bm{L}_\phi$ has the same sparsity pattern with $\bm{\Phi}$ 
and $\bm{G}$.

\section{Summary of the algorithm}
\label{sec_summary}

After evaluating sparse matrices, the probabilistic predictions can be
computed efficiently via Eq.~\eqref{eq_expectation} and \eqref{eq_uncer_aug}.
That is
\begin{equation} \label{eq_pred}
	\begin{split}
		\mu_\ast =& \bm{\phi}_\ast^\trans \bm{\omega}, \\
		\sigma_\ast^2 =& \sigma_n^2 \bm{\beta}^\top \bm{\beta}
		+ \left[ \phi^\ast(\bm{x}_\ast) \right]^2
	\end{split}
\end{equation}
with 
\begin{equation*}
	\bm{\omega} = \bm{\Phi}\bm{K}_y^{-1}\bm{y} = \sigma_n^{-2} \bm{\Phi}
	(\bm{I} - \bm{L}_\phi^\trans \bm{L}_\phi) \bm{y} \text{ and }
	\bm{\beta} = \bm{L}^{-1} \bm{\phi}_\ast.
\end{equation*}
In this work, the radial basis function $W_i(\bm{x})$ in the definition of 
$\phi^\ast(\bm{x})$ \eqref{eq_augbss} is chosen as the Wendland function 
with the support size the same with that of the basis function anchored 
on $\bm{x}_i$.

Notice that for each test point $\bm{x}_\ast$,
most entries in $\bm{\phi}_\ast = [\phi_i(\bm{x}_\ast]$ are zero since 
the basis functions are compactly supported.
Therefore, the prediction can be computed efficiently by only considering
basis functions that are non-zero on $\bm{x}_\ast$.
First we place $\bm{x}_\ast$ on nodes in the binary tree.
The node $\mathcal{N}^\ast_i$ is required in the prediction when
$\bm{x}_\ast$ lies inside its subregion.
Obviously at most $\mathcal{O}(\log n)$ nodes and $\mathcal{O}(m \log n)$
basis functions are required, thus the prediction 
can be evaluated in great efficiency.

The general algorithmic procedure is outlined in Algorithm \ref{alg_amrgp}.

\begin{algorithm}
	\caption{Adaptive Multi-Resolution Gaussian Processes}
	\label{alg_amrgp}
	\begin{algorithmic}[1]
		\Phase{Training}
			\Step{Construct the adaptive multi-resolution basis}
				\Substep Construct the adaptive binary tree
				\Substep Construct adaptive multi-resolution basis
				and shrink support domains when necessary
				\Substep Collect basis in the nested dissection order
			\Step{Evaluate matrices and vectors}
				\Substep Evaluate sparse matrices $\bm{\Phi}$ and $\bm{G}$.
				\Substep Compute the sparse Cholesky inverse for $\bm{G}$.
				\Substep Compute the sparse matrix $\bm{L}_\phi = \bm{L}^{-1} \bm{\Phi}$.
				\Substep Compute $\bm{\omega}$ and $\bm{\beta}$.
		\Phase{Prediction}
			\Step{Identify the nodes required in computing the 
				prediction on $\bm{x}_\ast$}
			\Step{Evaluate $\mu_\ast$ and $\sigma_\ast$ via 
				Eq.~\eqref{eq_pred}}
	\end{algorithmic}
\end{algorithm}

\section{Computational complexity analysis}
\label{sec_complexity}

Now let's analyse the complexity of the algorithm to see whether it is 
successfully brought down.
Since all the matrices are sparse and the matrix block sizes are bounded
by $m$, thus the key of the complexity analysis is to study the amount
of non-zero matrix blocks in the matrices.

Obviously the depth of the binary tree $L \sim \log_2 n$. 
There are about $2^l$ nodes on the $l$-th level.
Thus for a balanced binary tree, the total amount of nodes is
\begin{equation}
	\sum_{l=0}^L 2^l = 2^{L+1} - 1 \sim \mathcal{O}(2^{\log_2 n + 1} -1)
	\sim \mathcal{O}(n).
\end{equation}
For adaptive trees, the total amount of nodes should also be approximately 
of order $\mathcal{O}(n)$.

First analyse the complexity of evaluating $\bm{\Phi}$.
For each node at the $l$-th level, it has $l$ ancestors.
There are $\mathcal{O}(2^{-l} n)$ samples inside the node, thus it has 
$\mathcal{O}(2^{-l} n)$ descendants.
The amount of direct relatives of the node is
$\mathcal{O}(2^{-l} n) + l + 1$.
Notice $\bm{\Phi}(\mathscr{N}_i, \mathscr{N}_j)$ is non-zero 
only when $\mathscr{N}_i$ and $\mathscr{N}_j$ are relatives, thus
there are $\mathcal{O}(2^{-l} n) + l + 1$ non-zero blocks
corresponding to the node.
Therefore, the total amount of non-zero blocks 
$\bm{\Phi}(\mathscr{N}_i, \mathscr{N}_j)$ in $\bm{\Phi}$ is
\begin{equation}
	\begin{split}
		\# \bm{\Phi}(\mathscr{N}_i, \mathscr{N}_j)
		=& \sum_{l=0}^L 2^l \left( \mathcal{O}(2^{-l} n) + l + 1 \right) \\
		=& \sum_{l=0}^L \left[ \mathcal{O}(n) + \mathcal{O}(2^l l) + 2^l \right] \\
		\lesssim & \mathcal{O}(n \log n) + \sum_{l=0}^L \mathcal{O}(2^l L) + \mathcal{O}(n) \\
		\sim & \mathcal{O}(n \log n).
	\end{split}
\end{equation}
Since the size of these matrix blocks are bounded, thus the computational 
complexity of $\bm{\Phi}$ evaluation is $\mathcal{O}(n \log n)$.

Now let's study the complexity of computing 
$\bm{G} = \bm{\Phi} \bm{\Phi}^\trans + \sigma_n^2\bm{I}$.
Since $\bm{G}$ has the same sparsity pattern with $\bm{\Phi}$, thus 
there are also $\mathcal{O}(n \log n)$ non-zero blocks.
For each node $\mathscr{N}_i$ at the $l_i$-th level, the block 
$\bm{G}(\mathscr{N}_i, \mathscr{N}_j)$ is non-zero for 
$\mathscr{N}_j \in \mathcal{R}(\mathscr{N}_i)$, and
the matrix multiplication 
$\bm{\Phi}(\mathscr{N}_i, \mathscr{N}_k) \bm{\Phi}(\mathscr{N}_j, \mathscr{N}_k)$
has to be computed for each $\mathscr{N}_k$ with $\mathscr{N}_k \in 
\mathcal{R}(\mathscr{N}_i)$ and $\mathscr{N}_k \in \mathcal{R}(\mathscr{N}_j)$.
Assume $\mathscr{N}_j$ lies at the $l_j$-th level.
\begin{itemize}
	\item When $l_j <= l_i$, i.e., $\mathscr{N}_j \in \mathcal{A}(\mathscr{N}_i)
		\bigcup \{\mathscr{N}_i\}$, we have
		$\mathscr{N}_k \in \mathcal{R}(\mathscr{N}_i) 
		\subseteq \mathcal{R}(\mathscr{N}_j)$.
		Thus the amount of $\mathscr{N}_k$ is 
		$\sim \mathcal{O}(2^{-l_i n}) + l_i + 1$.
		There are one $\mathscr{N}_j$ for each $l_j$-th level.
	\item When $l_j > l_i$, i.e., $\mathscr{N}_j \in \mathcal{D}(\mathscr{N}_i)$,
		we have $\mathscr{N}_k \in \mathcal{R}(\mathscr{N}_j)
		\subseteq \mathcal{R}(\mathscr{N}_i)$.
		Thus the amount of $\mathscr{N}_k$ is
		$\sim \mathcal{O}(2^{-l_j n}) + l_j + 1$.
		There are $2^{l_j - l_i}$ $\mathscr{N}_j$'s for each $l_j$-th level.
\end{itemize}
Consequently, the total amount of matrix-matrix multiplications is
\begin{equation}
	\begin{split}
		& \sum_{l_i=0}^L \left\{
			\sum_{l_j=0}^{l_i} \left[ \mathcal{O}(2^{-l_i} n) + l_i + 1 \right] +
			\sum_{l_j=l_i+1}^L 2^{l_j-l_i} \left[
				\mathcal{O}(2^{-l_j} n) + l_j + 1 \right] \right\} \\
		=& \sum_{l_i=0}^L \left\{
			(l_i+1) \left[ \mathcal{O}(2^{-l_i} n) + l_i + 1 \right] +
			\sum_{l_j=l_i+1}^L \left[ \mathcal{O}(2^{-l_i} n) + (l_j+1) 2^{l_j-l_i} 
			\right] \right\} \\
		\lesssim & \sum_{l_i=0}^L \left[ (L+1) \mathcal{O}(2^{-l_i} n) + (l_i+1)^2 \right] + 
		\sum_{l_i=0}^L (L-l_i) \mathcal{O}(2^{-l_i} n) + 
		\sum_{l_i=0}^L (L+1) 2^{L-l_i+1} \\
		\lesssim & \mathcal{O}(n \log n) + \mathcal{O}(\log^3 n) + 
		\sum_{l_i=0}^L L\cdot \mathcal{O}(2^{-l_i} n) + (L+1) \mathcal{O}(n) \\
		\sim & \mathcal{O}(n \log n).
	\end{split}
\end{equation}
And the computational complexity for $\bm{G}$ evaluation is 
$\mathcal{O}(n \log n)$.

The complexity of sparse Cholesky inverse of a sparse matrix depends on the 
implementation. 
Generally, for a $n \times n$ sparse matrix $\bm{A}$ with $\# A_{ij}$ non-zero
entries, the sparse Cholesky inverse may be computed with
$\mathcal{O}(\# A_{ij} \log n)$
operations \citep{brandhorst2011sparse, liuwh1986cholesky}.
For the sparse Cholesky inverse of $\bm{G}$ in this work, since its 
size is $n$ and there are $\lesssim \mathcal{O}(n \log n)$ non-zero 
entries, thus the computational complexity is $\lesssim \mathcal{O}(n \log^2 n)$.

For the matrix-matrix multiplication $\bm{L}_\phi = \bm{L}^{-1} \bm{\Phi}$,
notice that $\bm{L}_\phi$ and $\bm{L}^{-1}$ has 
approximately the same sparsity pattern with $\bm{G}$ and $\bm{\Phi}$,
except that $\bm{L}^{-1}$ consists of only diagonal and sub-diagonal 
non-zero blocks.
Thus the complexity can be analysed in the similar manner as that for $\bm{G}$
evaluation, and the complexity is $\lesssim \mathcal{O}(n \log n)$.

Now let's study the complexity of the prediction process.
For a test point $\bm{x}_\ast$, 
we first identify the nodes required in the computation.
A node $\mathscr{N}^\ast$ is required only when $\bm{x}_\ast$ lies inside
the corresponding subregion.
If $\bm{x}_\ast$ lies inside the subregion of a leaf node, $\bm{x}_\ast$ also
lies inside the subregion of its ancestors.
Thus $\#\mathscr{N}^\ast \sim \mathcal{O}(\log n)$, and there are only 
$\mathcal{O}(\log n)$ non-zeros in $\bm{\phi}_\ast$.
Therefore, the computational complexity for the prediction 
$\mu_\ast = \bm{\phi}_\ast^\trans \bm{\omega}$ is $\mathcal{O}(\log n)$.

The predictive uncertainty consists of two parts, namely the aleatory uncertainty
$\sigma_n^2 \bm{\beta}^\trans \bm{\beta}$ and the epistemic uncertainty
$[\phi^\ast(\bm{x}_\ast)]^2$.
In the aleatory uncertainty evaluation, 
$\bm{\beta} = \bm{L}^{-1} \bm{\phi}_\ast$ has to be evaluated. 
Only $\mathcal{O}(\log n)$ columns of $\bm{L}^{-1}$ corresponding to non-zero
$\bm{\phi}_\ast$ are used in the computation.
Notice that each column of $\bm{L}^{-1}$ consists of at most $\mathcal{O}(\log n)$ 
non-zero values due to its sparsity pattern, thus the computational complexity 
for evaluating $\bm{\beta}$ is $\mathcal{O}(\log^2 n)$.

For the computation of the epistemic uncertainty $[\phi^\ast(\bm{x}_\ast)]^2$
with $\phi^\ast(\bm{x}) = \prod_{i=1}^n \left[ 1 - W_i (\bm{x}) \right]^\alpha$,
notice that $W_i$ in this work is chosen as the Wendland function anchored at
$\bm{x}_i$ with support size equal to the basis function $\phi_i(\bm{x})$.
Therefore, at most $\mathcal{O}(\log n)$ $W_i$ functions take non-zero values at 
a test point $\bm{x}_\ast$, thus the complexity for evaluating the epistemic 
uncertainty is $\mathcal{O}(\log n)$.
Consequently, the overall complexity for uncertainty computation is 
$\mathcal{O}(\log^2 n)$.

In summary, theoretically the computational complexity of our AMRGP is
$\mathcal{O}(n \log^2 n)$ for training and $\mathcal{O}(\log^2 n)$ for prediction.
The actual complexity depends on the implementation and 
may slightly differ from the theoretical prediction.

\section{Numerical experiments}
\label{sec_numexp}

In our AMRGP, the local basis functions in different nodes within the same level
have no intersecting support domains, thus they are independent and only 
correlated with their common ancestors.
This dependency allows us to develop a parallel implementation of our algorithm.
However, the parallelization is not the main contribution of this work,
thus our program is simply parallelized with OpenMP.

\begin{figure}[ht]
	\centering
	\subfigure[Case 1.] {
		\includegraphics[width=0.23\textwidth]{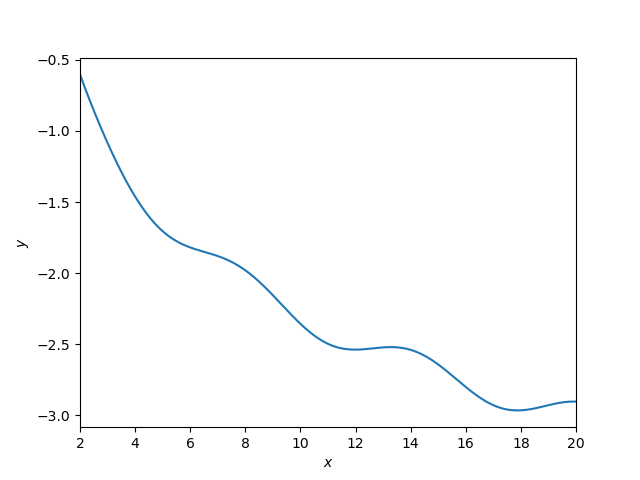}
	}
	\subfigure[Case 2.] {
		\includegraphics[width=0.23\textwidth]{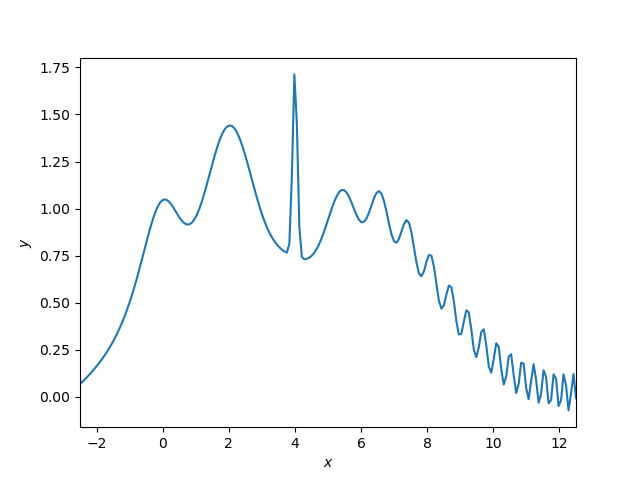}
	}
	\subfigure[Case 3.] {
		\includegraphics[width=0.23\textwidth]{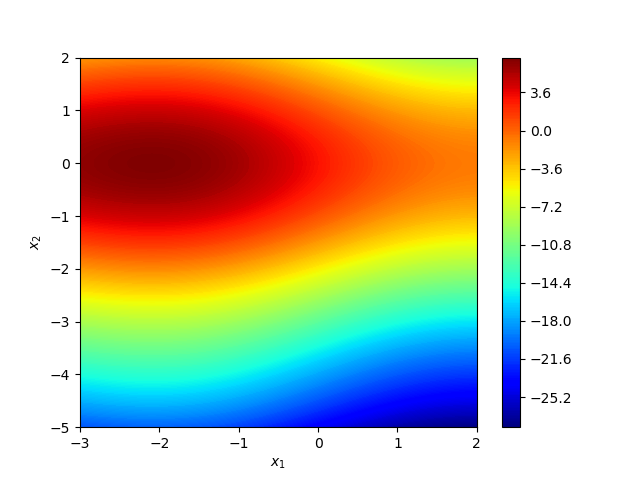}
	}
	\subfigure[Case 4.] {
		\includegraphics[width=0.23\textwidth]{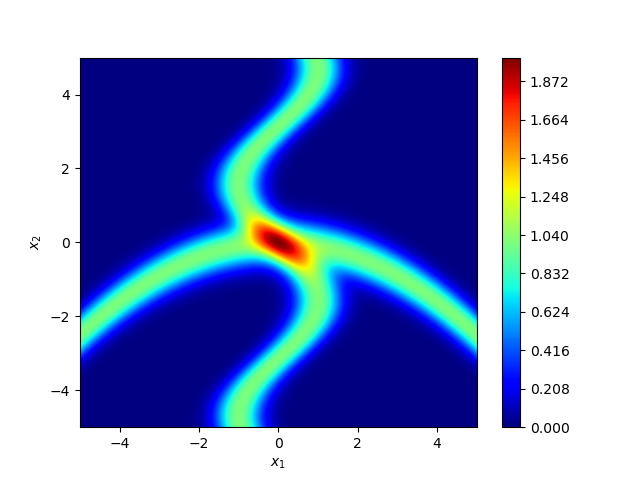}
	}
	\subfigure[Case 5 with $x_3=0$.] {
		\includegraphics[width=0.23\textwidth]{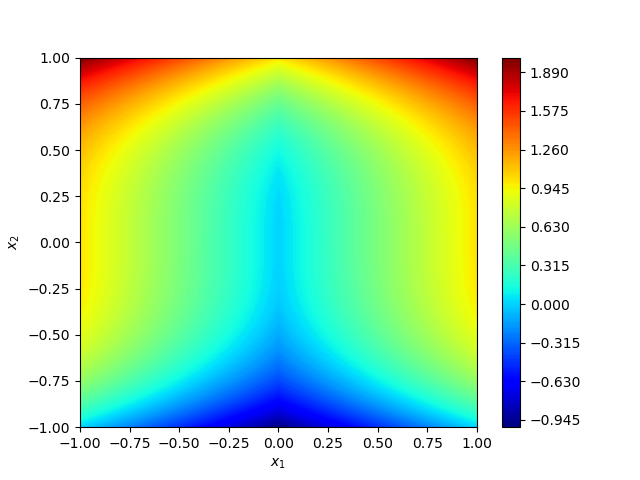}
	}
	\subfigure[Case 6 with $x_3=0$.] {
		\includegraphics[width=0.23\textwidth]{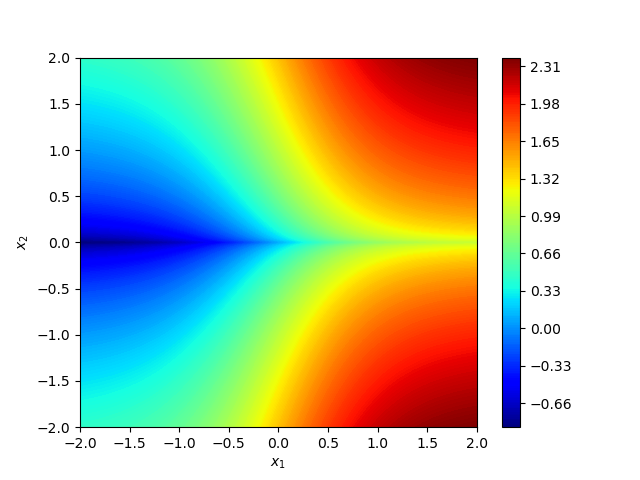}
	}
	\subfigure[Case 7 with $x_3=x_4=0$.] {
		\includegraphics[width=0.23\textwidth]{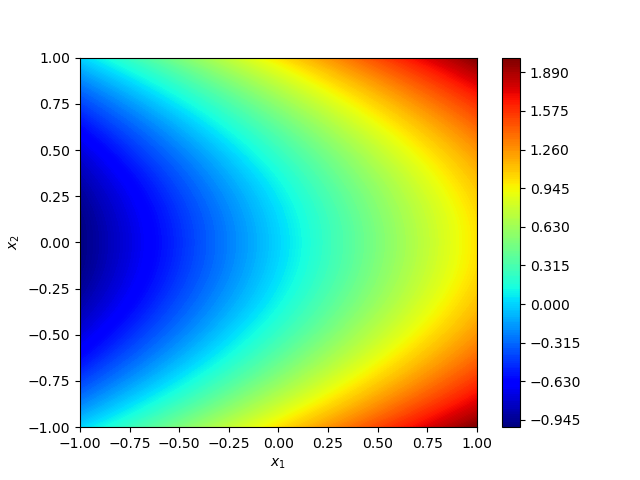}
	}
	\subfigure[Case 8 with $x_3=x_4=0$.] {
		\includegraphics[width=0.23\textwidth]{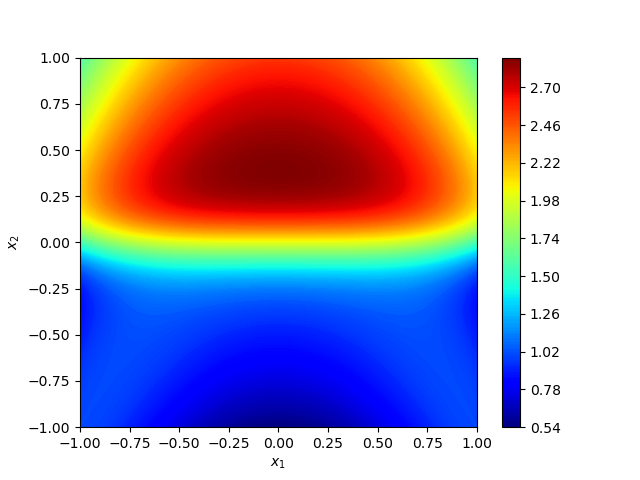}
	}
	\caption{Synthetic data.}
	\label{fig_synthetic}
\end{figure}

In this section, we study the performance of our AMRGP with numerical experiments.
Our program is implemented in C++ as an extension for Python.
All examples are carried out on CPUs at 2.90 GHz. 
In all numerical experiments except the one in Section \ref{subsec_physfld},
up to four-dimensional synthetic data in Table \ref{tab_synthetic} are tested.
For each dimension, both a smooth function and a function with relatively fast change
are considered, as illustrated in Fig.~\ref{fig_synthetic}.
Since we focus on high-fidelity modeling in this work, all training data are 
generated with a noise level of $\sigma_n^2 = 10^{-4}$.
This value is chosen as a trade-off between model fidelity and numerical robustness.
A lower noise level would further degrade the condition number of the covariance
matrix and lead to numerical instability, while a higher noise level would
compromise the high-fidelity requirement.

\begin{table} [ht]
	\centering
	\caption{Synthetic data used in numerical experiments.}
	\vspace{1em}
	\begin{tabular}{cccc}
		\hline
		Case	& $d$	& Function	& Range \\
		\hline
		1 & 1 & $-\log x_1 + 0.1 \sin x_1$ & [2, 20] \\
		2 & 1 & $\mathrm{e}^{-(x_1-2)^2} + \mathrm{e}^{-\frac{(x_1-6)^2}{10}} + \frac{1}{x_1^2+1} + \mathrm{e}^{-10 (x_1-4)^2} + 0.1 \sin 0.05 x_1^3$ & [-2.5, 12.5] \\
		3 & 2 & $-2 \sin x_1 + 3 \cos x_2 - x_1 - x_2^2$ & $[-3,2] \times [-5,2]$ \\
		4 & 2 & $\mathrm{e}^{-3(x_1 + \sin x_2)^2} + \mathrm{e}^{-3 (0.1 x_1^2 + x_2)^2}$ & $[-5,5]^2$ \\
		5 & 3 & $\abs{x_1} + x_2^3 + x_3^2$ & $[-1,1]^3$ \\
		6 & 3 & $x_1 \log(1+\abs{x_3}) + \tanh x_1 + \sqrt{\abs{x_2}}$ & $[-2,2]^3$ \\
		7 & 4 & $x_1 + x_2^2 + \log(1+x_3^2) + \sin x_4$ & $[-1,1]^4$ \\
		8 & 4 & $\cos(x_1^2+x_2) + \tanh 5 x_2 + 10 x_3 + \mathrm{e}^{-x_4^2}$ & $[-1,1]^4$ \\
		\hline
	\end{tabular}
	\label{tab_synthetic}
\end{table}

\subsection{Speedup of OpenMP parallelization}

\begin{figure} [ht]
	\centering
	\subfigure[Computational time cost.] {
		\includegraphics[width=0.45\textwidth]{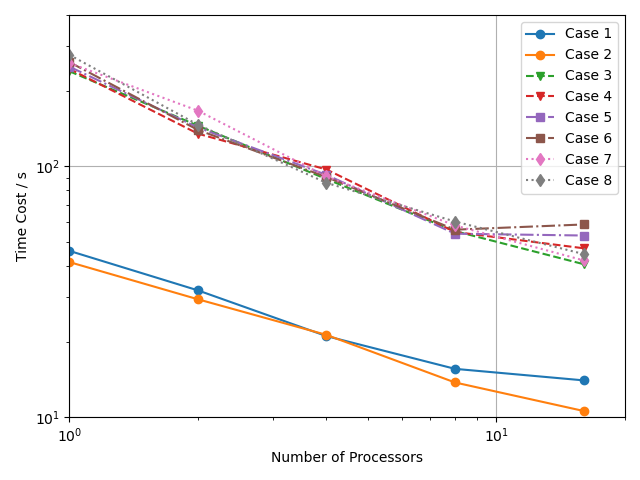}
	}
	\subfigure[Speedup.] {
		\includegraphics[width=0.45\textwidth]{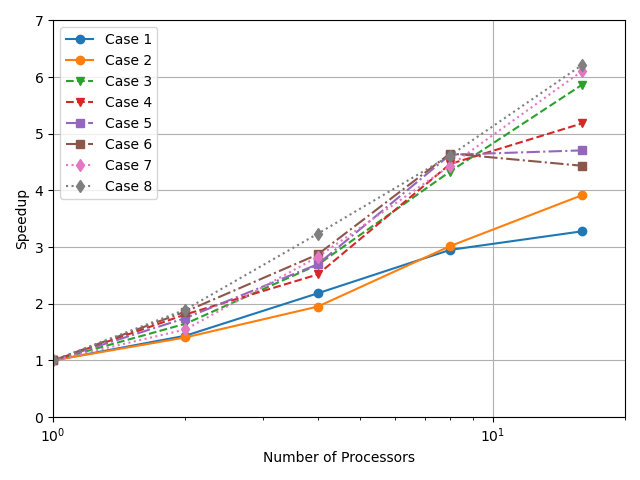}
	}
	\caption{OpenMP parallel computing with 1 million data points.}
	\label{fig_parallel}
\end{figure}

The speedup of our OpenMP parallelization is studied with $10^6$ training 
data points and $10^3$ predicting points for cases in Table \ref{tab_synthetic}.
Each case is computed multiple times with 1, 2, 4, 8, and 16 processors, 
respectively.
Parameters $\rho=2.5$ and $m=200$ are used for all the cases.
The computational time cost and speedup are illustrated in Fig.~\ref{fig_parallel}.

It is observed that the speedup ranges from 3 to 7 on 16 processors.
This is reasonable and consistent with typical limitations in shared-memory
parallel computing since our algorithm is memory-intensive with numerous
matrix blocks.
In a shared-memory system, all processors share the same memory bus and memory 
controller.
As more processors access memory simultaneously, contention increases, and the 
effective bandwidth per core decreases.
Thus the speedup for memory-intensive algorithms is often limited by the memory 
bandwidth rather than CPU compute power.
Although cache can help reduce the memory traffic, our AMRGP requires numerous
matrix blocks, leading to poor data locality and large working sets that exceed cache capacity. 
This forces frequent main memory accesses, amplifying the bandwidth bottleneck.

It is also shown that for one-dimensional cases, the time cost and speedup of 
parallel computing are typically lower than multi-dimensional cases.
This is because the number of local basis functions on one node increases 
exponentially with dimension until it reaches the limitation $m$.
For one-dimensional cases, the number of local basis functions
is typically much lower than $m$, leading to small matrix blocks and lower
computational cost.
The speedup is lower since typically the data locality is worse.

Parallelization with MPI would likely yield higher speedup and 
efficiency, but this is beyond of the scope of the present work and 
is left for future study.
Subsequent numerical experiments are computed in parallel with 16 processors
with OpenMP for efficiency unless otherwise stated.

\subsection{Choice of custom parameters}

There are two important parameters $\rho$ and $m$ in our AMRGP algorithm.
Here $\rho$ is the scaling factor of the basis support sizes, and $m$ is
the upper bound on the number of local basis functions on each node 
as well as the corresponding matrix block size.
Clearly, when $\rho$ and $m$ tend to infinity, there would be very few or even no 
local basis functions on non-root nodes, 
and nearly all local basis functions lie on the root, thus the computational
complexity of the GP regression cannot be reduced.
Therefore, we suggest defining these two parameters manually rather than
through optimization to ensure computational efficiency.

Theoretically, for a fixed $\rho$, there tend to be fewer data points inside
the support region of each local basis function in high dimensions, making
it difficult to capture global trends.
Therefore, for high-dimensional cases,
$\rho$ should take a relatively large value.
On the other hand, the matrix block size increases with $\rho$ and grows
exponentially with dimension, so for high-dimensional cases, it must 
be bounded by $m$ to ensure efficiency.
Hence, the trade-off between accuracy and efficiency becomes more critical
in high dimensions, requiring careful selection of $\rho$ and $m$.
In this section, we study the choice of $\rho$ and $m$ using 
four-dimensional cases only, and then validate the chosen parameters with other cases.

\begin{figure}[ht]
	\centering
	\subfigure[CPU time cost.] {
		\includegraphics[width=0.31\textwidth]{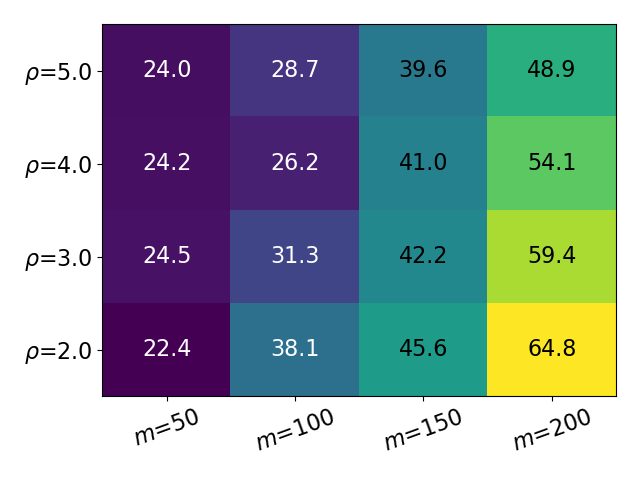}
	}
	\subfigure[Memory cost.] {
		\includegraphics[width=0.31\textwidth]{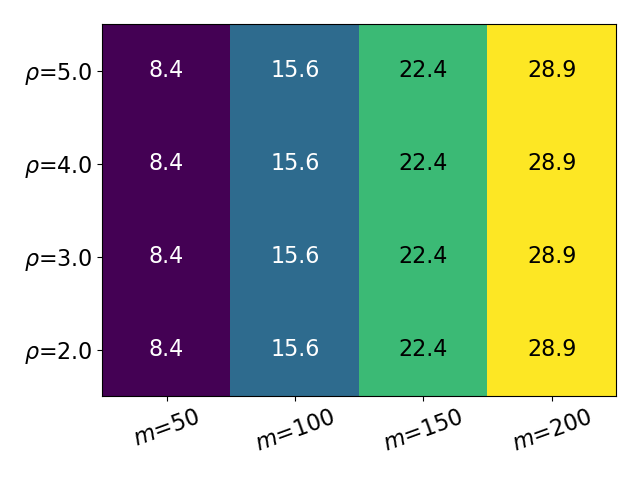}
	}
	\subfigure[Normalized CRPS.] {
		\includegraphics[width=0.31\textwidth]{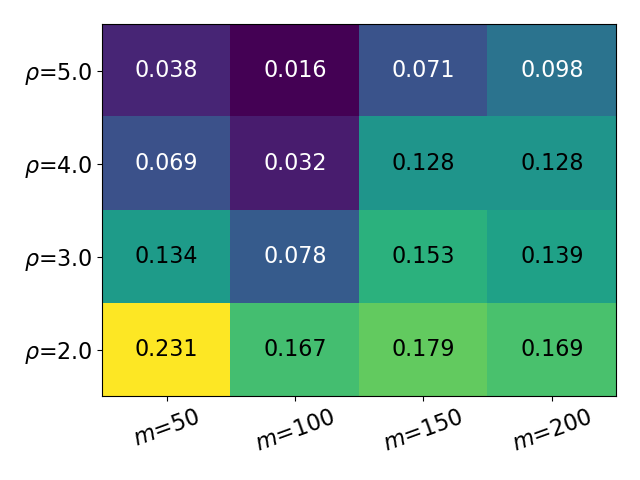}
	}
	\caption{Performance of our AMRGP on Case 7 with different $\rho$'s and $m$'s.}
	\label{fig_case4d1_rhom}
\end{figure}

\begin{figure}[ht]
	\centering
	\subfigure[CPU time cost.] {
		\includegraphics[width=0.31\textwidth]{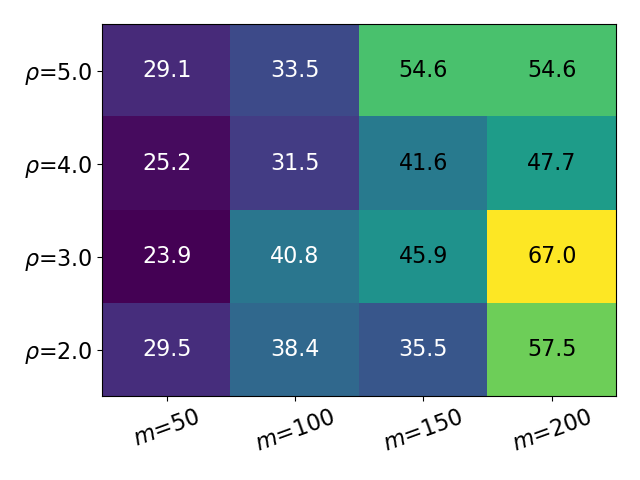}
	}
	\subfigure[Memory cost.] {
		\includegraphics[width=0.31\textwidth]{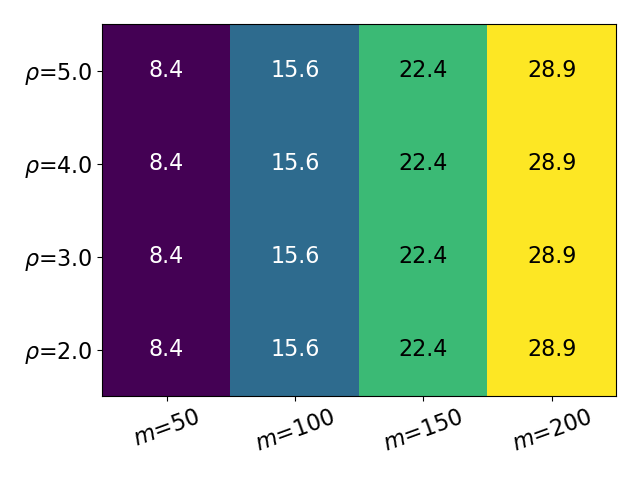}
	}
	\subfigure[Normalized CRPS.] {
		\includegraphics[width=0.31\textwidth]{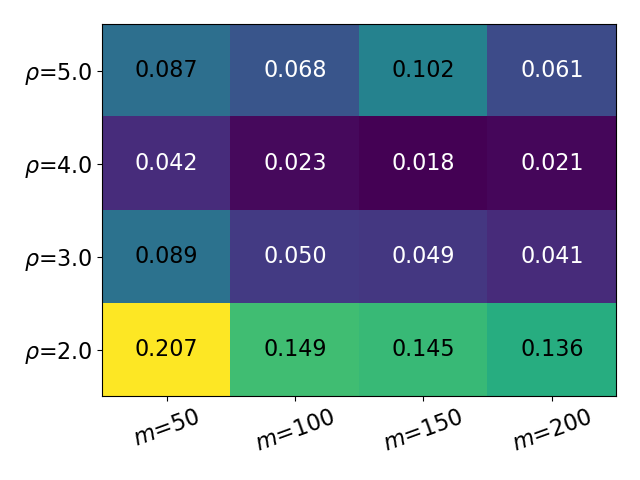}
	}
	\caption{Performance of our AMRGP on Case 8 with different $\rho$'s and $m$'s.}
	\label{fig_case4d2_rhom}
\end{figure}

The time cost $T$, memory cost $M$, and normalized CRPS for four-dimensional cases
are depicted in Fig.~\ref{fig_case4d1_rhom} and \ref{fig_case4d2_rhom}.
It is shown that the computational cost generally increases with $m$, and
is less sensitive to $\rho$. This is reasonable since for four-dimensional 
cases, the number of local basis functions increases to the upper bound $m$ quickly,
thus the matrix block sizes are generally determined by $m$.
The normalized CRPS, however, is mainly influenced by $\rho$.
For Case 7, the normalized CRPS generally decreases with $\rho$, indicating
that better predictions can be achieved with larger $\rho$.
For Case 8, however, the normalized CRPS first decreases with $\rho < 4.0$ 
and then increases.
This is because an overly small $\rho$ leads to overly small support sizes 
of basis functions that produce insufficiently smooth predictions,
but an overly large $\rho$ may results in difficulty in capturing
fine-scale features.
Thus theoretically $\rho$ should take larger values for smooth functions like Case 7 and 
relatively smaller values for functions with fast changes like Case 8.
Larger $m$ generally leads to better predictions, especially for Case 8.
However, too large $m$ leads to poorer predictions in Case 7.
The reason is that larger $m$ reduces the number of shrunk support regions, 
leading to higher collinearity among columns of $\bm{\Phi}$ and a worse condition
number of $\bm{K}_y$, which greatly amplifies round-off errors.
Therefore, the best choice of $\rho$ and $m$ may depend on cases.
For the above two cases, $\rho=4.0$ and $m=100$ gives a good balance between
efficiency and accuracy, which may provide a good starting point in adjusting
parameters for general cases.

Now we validate the choice $\rho=4.0$ and $m=100$ with one- to three-dimensional cases.
The computational cost and normalized CRPS's are listed in Table~\ref{tab_valid_rhom}.
Relatively good predictions can be achieved except Case 3.
For this case, by slightly adjusting $\rho$ to 3.0, the normalized CRPS becomes
0.035. Therefore, we suggest take $\rho=4.0$ and $m=100$ as
the initial choice for general cases, then adjust the values slightly
if the prediction is unsatisfactory.
In the rest of this work, $\rho=4.0$ and $m=100$ are taken unless otherwise stated. 

\begin{table} [ht]
	\centering
	\caption{Performance of the choice $\rho=4.0$ and $m=100$ on Case 1--6 with 1 million training data points.}
	\vspace{1em}
	\begin{tabular}{c|cccccc}
		\hline
		Case & 1 & 2 & 3 & 4 & 5 & 6 \\
		\hline
		Time (s)		&  25.9 &  20.5 &  19.9 &  23.4 &  20.6 &  20.2 \\
		Memory (GB)		&  11.5 &  11.5 &  11.5 &  11.5 &  11.5 &  11.5 \\
		Normalized CRPS	& 0.004 & 0.002 & 0.345 & 0.009 & 0.013 & 0.024 \\
		\hline
	\end{tabular}
	\label{tab_valid_rhom}
\end{table}

\subsection{Computational complexity}

\begin{figure}[ht]
	\centering
	\subfigure[Training time cost.] {
		\includegraphics[width=0.31\textwidth]{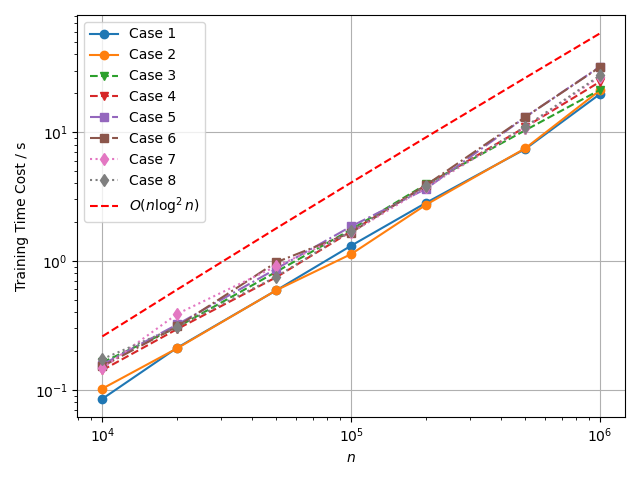}
	}
	\subfigure[Prediction time cost.] {
		\includegraphics[width=0.31\textwidth]{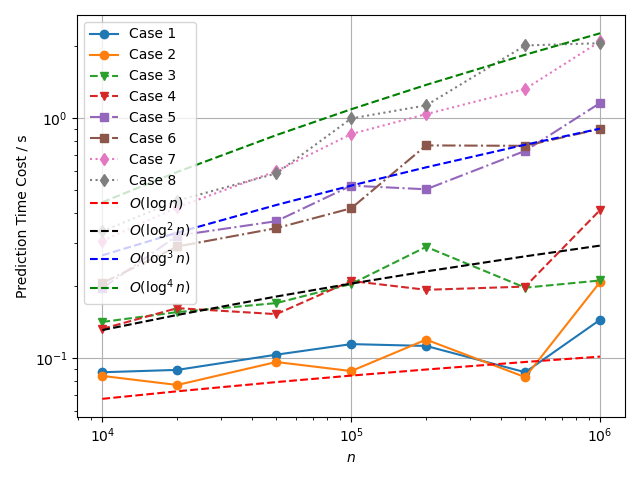}
	}
	\subfigure[Memory cost.] {
		\includegraphics[width=0.31\textwidth]{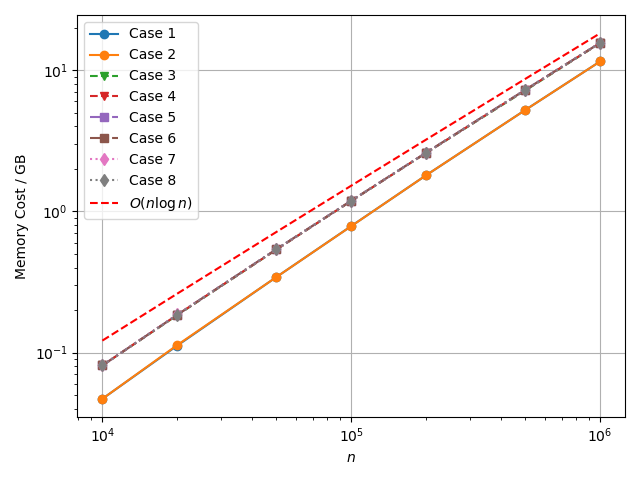}
	}
	\caption{Complexity of our AMRGP.}
	\label{fig_complexity}
\end{figure}

The computational complexity of our AMRGP is analysed theoretically in Section
\ref{sec_complexity}. However, the actual complexity depends on the implementation
and may differ from the theoretical prediction.
Thus here we study its actual computational complexity via numerical experiments.
The cases in Table \ref{tab_synthetic} are computed for $n$ ranging from 
$10^4$ to $10^6$.
The training time, prediction time and memory cost are illustrated in 
Fig.~\ref{fig_complexity}.
It is shown that the training time and memory cost increase approximately 
as $\mathcal{O}(n \log^2 n)$ and $\mathcal{O}(n \log n)$, respectively, which
is consistent with the theoretical prediction.
The prediction time scales approximately as $\mathcal{O}(\log^d n)$ with $d$ 
being the dimensionality, which differs somewhat from the theoretical
prediction of $\mathcal{O}(\log^2 n)$. 
Overall, the computational complexity of our AMRGP is about
$\mathcal{O}(n \log^2 n)$ in training and $\mathcal{O}(\log^d n)$ in prediction.

\subsection{Comparison to other scalable GP models}

We now compare our AMRGP with other scalable GP models.
Among various scalable GPs, we mainly take three representative models as follows
to make comparison by computing the same cases.
\begin{enumerate}
	\item Stochastic variational GP (SVGP) \cite{hensman2013gp}:
		A well known scalable GP model providing good low rank approximations
		for GPs by combining the variational inference technique with the idea of 
		inducing variables.
		It has been implemented in GPyTorch.
	\item Structured kernel interpolation for products (SKIP) \cite{gardner2018skip}:
		An efficient extension of grid-based SKI to multi-dimensional cases,
		which is highly efficient and has been implemented in GPyTorch.
	\item Scaled Vecchia GP (SVecGP) \cite{katzfuss2022svecgp}:
		A recently developed scalable GP model based on the Vecchia approximation. 
		It is part of the R package \texttt{GPVecchia} which is a general 
		framework for various Vecchia approximations.
\end{enumerate}
Note that our AMRGP is a scalable exact GP with a multi-resolution structure,
making it closely related to Noack's scalable exact GP \cite{noack2023nonstationary} 
and the highly optimized MRA for GP \cite{huanghuang2019mrgp}.
These two models, however, are not compared directly in this manner, since 
we encountered difficulties in implementing them on our platform.
Instead, a preliminary comparison will be provided by comparing the reported 
time costs in the literature.

In standard GP model and its approximations including SVGP and SKIP, radial basis 
functions and Mat\'ern's kernels are commonly used.
Generally length scales and other hyper-parameters are optimized 
to match the smoothness of the target function.
In contrast, our AMRGP employs multi-resolution basis functions that can adapt
to different scales without explicit length scale optimization.
However, this approach may be less effective in capturing different rates of 
changes along different directions in multi-dimensional problems, as it lacks
the ability to tune per-dimension length scales.
In this comparison, we do not perform hyper-parameter optimization in our 
AMRGP, while other scalable GPs are trained with optimization.
This may lead AMRGP yielding less accurate predictions than the optimized 
models in certain scenarios.

It should be noted that a fair comparison between these models is challenging.
SVGP and SKIP are compute-intensive and can be greatly 
accelerated by GPU computations.
SVecGP and our AMRGP, however, are memory-intensive and thus more 
suitable for parallel computing with multiple processors.
In this work, all models are computed using 16 processors in parallel
without GPU acceleration. 
Moreover, the optimal settings for these GP models are beyond the scope of this research.
The choice of number of inducing variables in SVGP and grid points in SKIP $m$
may range from 100 to 1,000 in literature \cite{hensman2013gp, gardner2018skip, 
wuluhuan2022vnngp}.
In this work, we choose $m=200$ as a moderate value that balances accuracy and
computational cost.
The size of the conditioning set in SVecGP is set as $m=20$, which is the 
choice used in the numerical example of the \texttt{GPVecchia} package.
Both SVGP and SKIP are optimized with 50 iterations via the Adam optimizer,
with learning rates of 0.1 and 0.05, respectively.
The number of iterations of SVecGP is controlled by a convergence criterion.
We found that only a few iterations can provide quite accurate predictions, thus
we set the tolerance to 1.0 which typically requires fewer than 10 iterations
to converge.
These scalable GPs are compared using $n=100,000$ training data points
for the synthetic cases in Table~\ref{tab_synthetic}, and 1,000 
test points are used to compute the predictive accuracy.

\begin{figure}[ht]
	\centering
	\includegraphics[width=0.9\textwidth]{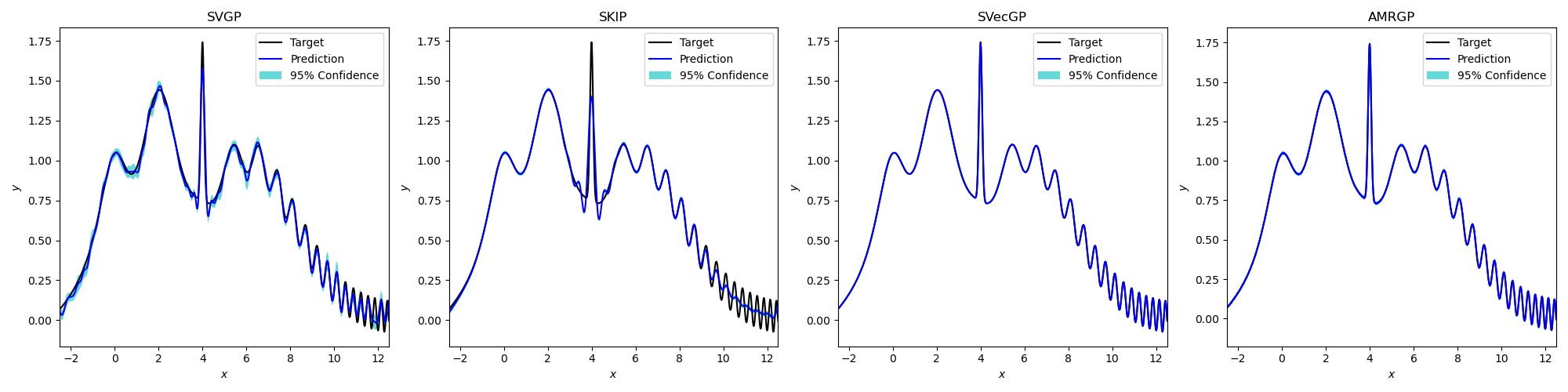}
	\caption{Predictions for Case 2.}
	\label{fig_case2}
\end{figure}

\begin{figure}[ht]
	\centering
	\includegraphics[width=0.9\textwidth]{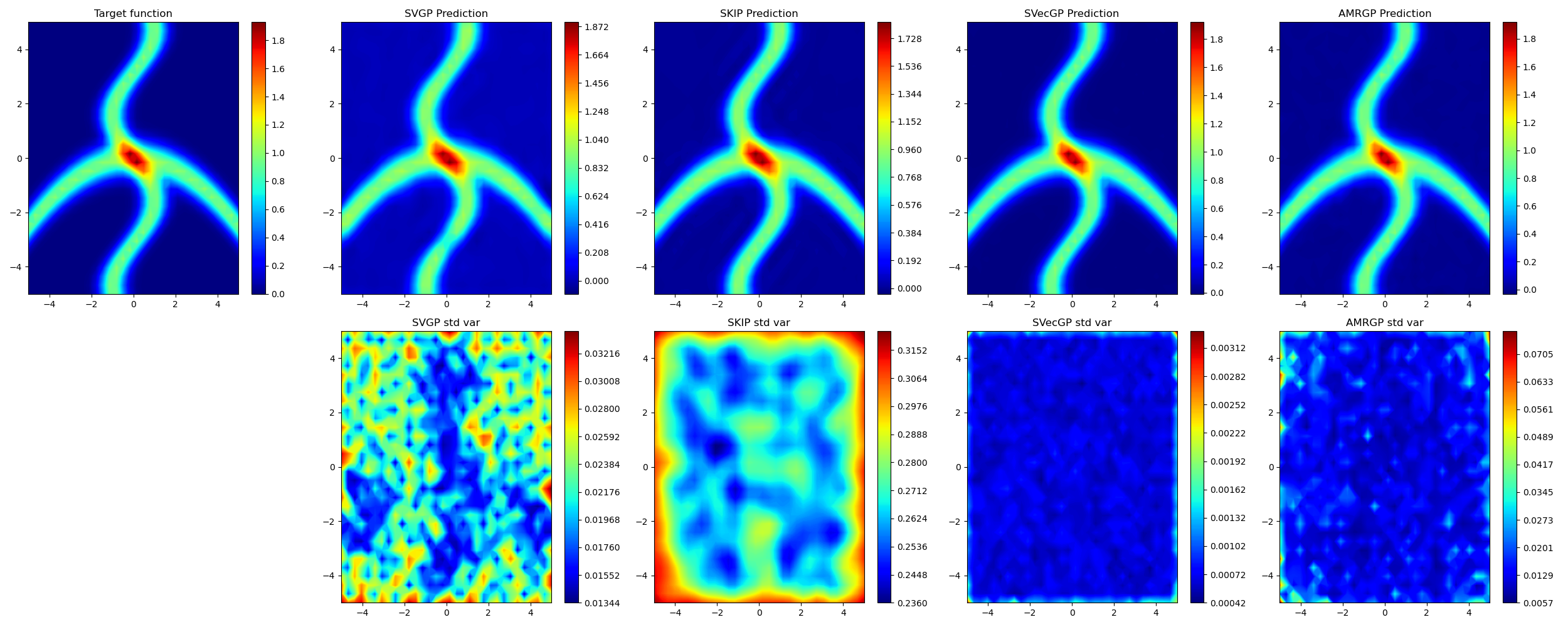}
	\caption{Predictions for Case 4: exact and predicted values (top) and standard deviations (bottom).}
	\label{fig_case4}
\end{figure}

The computational cost and normalized CRPS of these scalable GP models are listed
in Table~\ref{tab_cmpgps}.
It is shown that our AMRGP provides predictions competitive with
SVGP and SKIP, and in some cases outperforms them.
Case 2 and 4, where AMRGP and SVecGP outperform SVGP and SKIP, 
are illustrated in Fig.~\ref{fig_case2} and \ref{fig_case4}.
Both of these two cases are functions with multi-scale features.
SVGP and SKIP have difficulty in capturing the fine-scale fluctuations in Case 2
and predicting the peak value in Case 4.
In such cases, both SVecGP and our AMRGP provide better predictions.
This is as expected since both SVecGP and AMRGP have hierarchical structures, 
which are more suitable for capturing the multi-resolution features.

\begin{table} [ht]
	\centering
	\caption{Comparison of scalable GP models. The time in parenthesis is the 
	average time cost by one iteration in approximate GPs and the training time
	in AMRGP.}
	\vspace{1em}
	\begin{tabular}{ccccccc}
		\hline
		Case	& Dimension & Metric	& SVGP	& SKIP	& SVecGP & \bf{AMRGP} \\
		\hline
		\multirow{2}{*}{1}	&\multirow{2}{*}{1}	& Time cost (s)	    & 123.4 (2.5)  &   77.7 (1.5) & 58.8 (7.1)  & \bf{5.1 (3.8)}   \\
				&					& Normalized CRPS $\times$ 100	& 0.12   &   0.02 & 0.01  & \bf{0.10}  \\
		\hdashline
		\multirow{2}{*}{2}	&\multirow{2}{*}{1}	& Time cost (s)		& 125.3 (2.5)  &   86.1 (1.5) & 57.4 (6.9)  & \bf{3.3 (1.6)}   \\
				&					& Normalized CRPS $\times$ 100	& 2.11   &   2.73 & 0.07  & \bf{0.21}  \\
		\hdashline
		\multirow{2}{*}{3}	&\multirow{2}{*}{2}	& Time cost (s)		& 131.7 (2.6) &  176.0 (3.4) & 84.0 (8.9) & \bf{2.8 (2.3)}   \\
				&					& Normalized CRPS $\times$ 100	& 0.22   &   0.59 & 1.15  & \bf{0.99}  \\
		\hdashline
		\multirow{2}{*}{4}	&\multirow{2}{*}{2}	& Time cost (s)		& 126.8 (2.5) &  173.1 (3.4) & 82.1 (8.9) & \bf{2.7 (1.8)}   \\
				&					& Normalized CRPS $\times$ 100	& 3.42   &  16.07 & 0.57  & \bf{1.60}  \\
		\hdashline
		\multirow{2}{*}{5}	&\multirow{2}{*}{3}	& Time cost (s)		& 131.6 (2.6) &  258.5 (5.0) & 102.7 (9.6) & \bf{3.9 (2.1) }   \\
				&					& Normalized CRPS $\times$ 100	& 1.64   &   3.52 & 0.64  & \bf{1.46}  \\
		\hdashline
		\multirow{2}{*}{6}	&\multirow{2}{*}{3}	& Time cost (s)		& 133.2 (2.7) &  258.6 (5.0) & 161.5 (9.4) & \bf{3.6 (1.9)}   \\
				&					& Normalized CRPS $\times$ 100	& 2.62   &   5.98 & 1.17  & \bf{3.18}  \\
		\hdashline
		\multirow{2}{*}{7}	&\multirow{2}{*}{4}	& Time cost (s)		& 138.4 (2.8) &  305.2 (5.9) & 258.4 (4.6) & \bf{4.9 (3.0)}   \\
				&					& Normalized CRPS $\times$ 100	& 3.52   &   1.26 & 0.29  & \bf{1.53}  \\
		\hdashline
		\multirow{2}{*}{8}	&\multirow{2}{*}{4}	& Time cost (s)		& 146.5 (2.9) &  308.8 (5.8) &  98.2 (12.0) & \bf{5.1 (2.2)}   \\
				&					& Normalized CRPS $\times$ 100	& 2.40   &   1.04 & 8.16  & \bf{2.31}  \\
		\hline
	\end{tabular}
	\label{tab_cmpgps}
\end{table}

Compared to SVecGP, AMRGP is slightly less accurate for most cases
except for Case 3 and 8.
This is reasonable since SVecGP uses an anisotropic kernel function with 
optimized length scales to fit the smoothness along different directions, 
while the basis functions in AMRGP are isotropic.
It should be noted again that although our AMRGP is an exact GP model while
SVecGP is an approximate one, SVecGP can still yield better predictions 
because of its kernel flexibility.
The unusally high normalized CRPS of SVecGP in Case 8 may be attributed to the
insufficient hyper-parameter optimization for rapidly varying functions in 
high dimensions.
This comparison also highlights that an exact GP model does not necessarily 
outperform approximate ones, especially when the latter employ more flexible
kernel structures.

The reported time cost in Table~\ref{tab_cmpgps} includes both training and 
prediction overhead. The value in parenthesis is the average time cost 
by one iteration in approximate GPs and the training time in AMRGP, which
are approximately the time cost by computing a single covariance matrix inverse.
It is shown that 
for all these cases, our AMRGP is over 10 times faster than all the other three 
scalable GP models.
The efficiency gain of AMRGP mainly stems from its construction:
the GP model is directly built from multi-resolution basis functions, which 
can capture multi-scale features even without iterative 
hyper-parameter optimization.
For a single covariance matrix inverse,
our AMRGP achieves efficiency comparable with or even better than SVGP
and SKIP, and always significantly outperforms SVecGP.

In addition to the above direct comparisons, we also provide preliminary 
comparisons with Noack's scalable exact GP \cite{noack2023nonstationary} and 
the highly optimized MRA for GP \cite{huanghuang2019mrgp}. 
We only compare with the reported time costs rather than by computing the same 
cases on the same platform, since we encountered difficulties in implementing 
them on our platform.

It is reported that for 5 million two-dimensional data points, Noack's scalable
exact GP model takes over 24 hours with 256 GPUs.
In contrast, for 5 million two-dimensional data points generated from Cases 3 
and 4, our AMRGP takes only 182 and 122 seconds with 16 CPUs, respectively.
That is, even without GPU acceleration, our AMRGP is over 500 times faster 
than Noack's scalable exact GP in terms of training time.

The highly optimized MRA for GP is parallelized with MPI, while our AMRGP is 
parallelized with OpenMP.
To eliminate the influence of different parallelization strategies, we compare 
their efficiency in serial computation using only one processor.
For approximately 2.44 million two-dimensional data points, the number of 
knots per node in MRA is set to 49 to reduce memory cost.
It then takes about 154.63 seconds to evaluate the log marginal likelihood 
once, i.e., a single covariance matrix inversion. 
To make a relatively fair comparison, the upper bound of the number of local 
basis functions in our AMRGP is set to 50.
With 2.44 million two-dimensional data points generated from Cases 3 and 4, our 
AMRGP takes 187 and 175 seconds, respectively, which is about 13--21\% slower 
than the highly optimized MRA.
It is worth noting again that these comparisons are preliminary due to 
differences in hardware and implementation.

\subsection{Validation with physical fields}
\label{subsec_physfld}

Recently GPs has been applied in solving partial differential equations 
\cite{raissi2017ml, raissi2018numerical, wangjunyang2021bayesian, owhadi2023gph,
guohongwei2026pigp}.
Modeling physical fields with GP regression is one of the key of the GP-based
numerical solver.
Thus here we validate our AMRGP via modelling physical fields.
Although the following test cases have analytical solutions, they represent
physically meaningful fields with specific smoothness and multi-scale features.
This validate AMRGP's ability to model realistic physical fields, which is
a prerequisite for GP-based PDE solvers. 
The training data are sampled from the analytical solution, mimicking the
scenario where observations are available at discrete points.

First consider the numerical case with homogeneous Burger's equation in 
\cite{wangjunyang2021bayesian} which describes a shock formation process:
\begin{equation}
	\frac{\partial u}{\partial t} + u \frac{\partial u}{\partial x} 
	- \alpha \frac{\partial^2 u}{\partial x^2} = 0, \quad t \in [0, T], x \in [0, L].
\end{equation}
The initial and boundary conditions are defined by
\begin{subequations}
	\begin{equation*}
		u(0, x) = 2 \alpha \left( \frac{a k \sin(kx)}{b + a \cos(kx)} \right), 
		\quad x \in [0, L],
	\end{equation*}
	\begin{equation*}
		u(t, 0) = u(t, L) = 0,	\quad t \in [0, T],
	\end{equation*}
\end{subequations}
and for this experiment, $\alpha = 0.02, a = 1, b = 2, k = 1, T=30$ and $L=2\pi$.
These initial and boundary conditions gives a closed-form analytical solution
\begin{equation}
	u(t,x) = 2\alpha \left( \frac{a k \exp(-\alpha k^2 t) \sin(kx)}{b + a \exp(-\alpha k^2 t) \cos(kx)} \right).
\end{equation}

\begin{figure}[ht]
	\centering
	\includegraphics[width=0.9\textwidth]{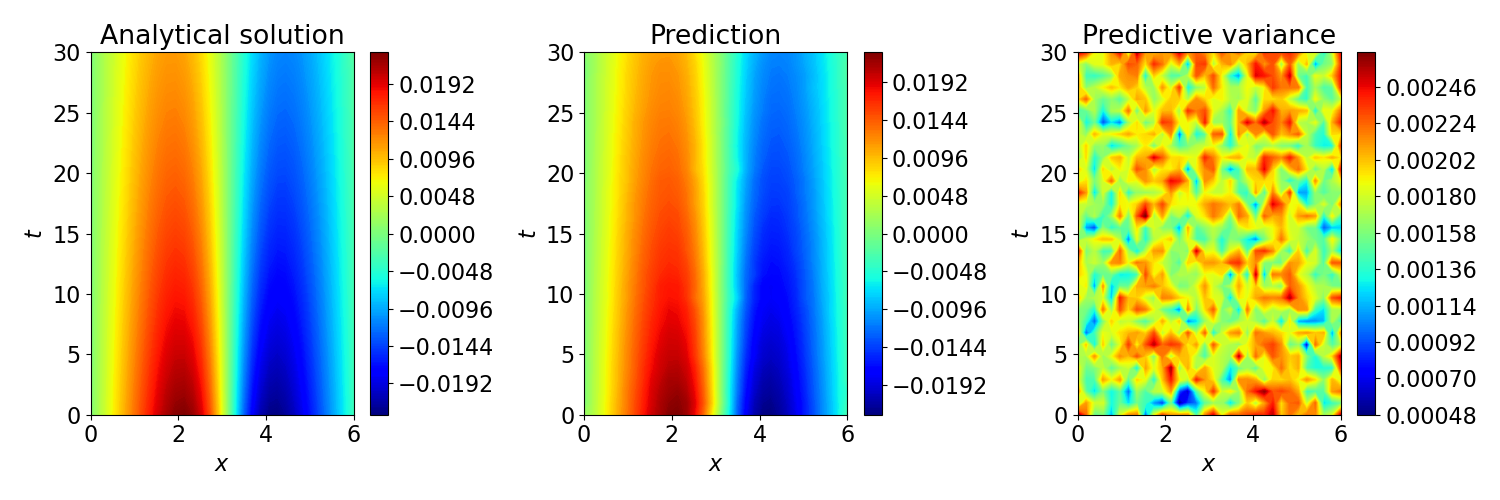}
	\caption{Prediction for the Burger's equation over the computational domain.}
	\label{fig_burgers}
\end{figure}

We test our AMRGP with six million training data points sampled in the 
computational domain, and then test its accuracy with 1,000 test points.
It takes 107.5 GB memory, 174 seconds for training and 1.1 seconds for 
prediction, and the normalized CRPS is about 0.04.
The prediction is compared with the analytical solution
in Fig.~\ref{fig_burgers}.

The second physical fields used to validate our AMRGP is oscillatory and is 
governed by the wave equation in \cite{raissi2018numerical}:
\begin{equation}
	\frac{\partial^2 u}{\partial t^2} = \frac{\partial^2 u}{\partial x^2}
\end{equation}
The initial and boundary conditions are defined by
\begin{subequations}
	\begin{equation*}
		u(0, x) = \frac{1}{2} \sin (\pi x), \quad
		\frac{\partial u}{\partial t} = \pi \sin (3 \pi x),
	\end{equation*}
	\begin{equation*}
		u(t, 0) = u(t, 1) = 0.
	\end{equation*}
\end{subequations}
The analytical solution is 
\begin{equation}
	u(t, x) = \frac{1}{2} \sin(\pi x) cos(\pi t) + \frac{1}{3} \sin(3\pi x) \sin(3 \pi t).
\end{equation}

\begin{figure}[ht]
	\centering
	\includegraphics[width=0.9\textwidth]{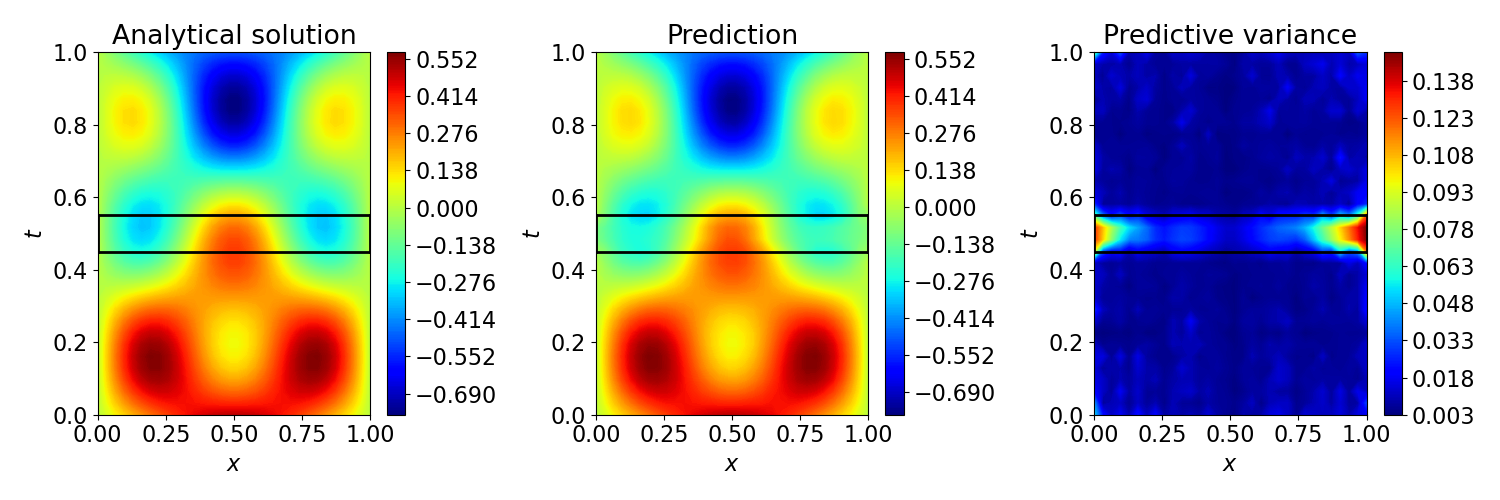}
	\caption{Predicted for the wave equation over the computational domain.
	The black rectangle denotes the test region.}
	\label{fig_wave}
\end{figure}

We test our AMRGP with the case that the test points lie outside the training
region. Here we divide the computational domain $[0,1] \times [0,1]$ into the
training region $[0, 1] \times [0, 0.45] \bigcup [0, 1] \times [0.55, 1]$,
and leaves its complement $[0, 1] \times [0.45, 0.55]$ as the test region.
We train our AMRGP model with six million training data taken from the
training region, then test the prediction with 1,000 points in the test region.
It takes 109.6 GB memory, 178 seconds for training and 0.14 seconds 
for prediction, and the normalized CRPS is 0.11.
The prediction time is much less than that for the Burger's equation case.
This is because, when the test points lie outside of the training region,
its prediction is computed only with coarse-scale basis functions on
the high levels of the binary tree.
The comparison between the prediction and the analytical solution is shown
in Fig.~\ref{fig_wave}.
The local minima in the test region are not correctly predicted since 
only large-scale trends are used in the prediction when the test points
lies outside of the training region.
The predictive variance in the test region is much higher than
that in the computational region due to the high epistemic uncertainty.

\section{Conclusion}
\label{sec_conclusion}

A scalable exact GP framework with adaptive multi-resolution basis is proposed
in this work, which bridges the gap between approximation-based scalability 
and high-fidelity modeling. 
Unlike predominant scalable GPs that sacrifice exact inference for efficiency, 
our AMRGP constructs a naturally data-sparse model that enables exact, log-linear 
computation.

The key innovations are summarized as follows.
\begin{itemize}
	\item An exact, data-sparse representation of the covariance matrix 
		constructed from compactly supported multi-resolution basis functions.
	\item A mathematically precise and efficient computational scheme 
		for computing the inverse of the data-sparse covariance matrix using
		the Sherman-Morrison-Woodbury formula and the sparse Cholesky inverse 
		algorithm.
	\item An augmentation technique for modeling the epistemic uncertainty 
		in GPs with compactly supported basis functions.
	\item Multi-resolution basis functions anchored directly to samples, with
		support sizes varying smoothly from the finest to the coarsest scale.
		This circumvents the need for complicated selection or optimization of
		auxiliary points, and
		theoretically enables AMRGP to adapt to arbitrarily distributed data
		and capture features at any scale.
	\item A support domain shrinkage technique that limits non-zero matrix block 
		sizes for multi-dimensional cases, enabling $\mathcal{O}(n \log^2 n)$
		computational complexity for training and $\mathcal{O}(\log^d n)$
		for prediction.
\end{itemize}

By achieving scalability through exact inference with a data-sparse model,
rather than approximating a dense model, our AMRGP provides a more
principled foundation for large-scale non-parametric regression.
Numerical experiments show that the efficiency of our AMRGP is 
comparable with or even better than existing scalable approximate GPs.

It should be noted that currently no hyper-parameter optimization is performed 
in our AMRGP. This may limit its effectiveness for multi-dimensional 
functions with anisotropic variations.
Moreover, the epistemic uncertainty is given directly by the augmented basis
function, which depends only on training data locations.
This may lead to inaccurate predictive variance.
These imperfections may be addressed by incorporating automatic relevance
determination (ARD) and uncertainty calibration techniques.
In addition, more advanced techniques should be investigated
to mitigate the ill-conditioning of the covariance matrix and 
enhance numerical robustness while preserving high model fidelity.
These issues are beyond the scope of this work and left for future study.

\section*{Acknowledgements}

The authors sincerely appreciate the support and discussions provided by all
the members of Prof. Heng Yong's AI++ team.
This work is partially supported by the National Natural Science Foundation of 
China (NSFC) under Grant Nos. 12331010 and 12101064,
the Presidential Foundation of China Academy of 
Engineering Physics Project under Grant No. YZJJZQ2024018, and
the National Safety Academic Fund (NSAF) under Grant No. U2230208.

\section*{Bibliography}
\addcontentsline{toc}{chapter}{Bibliography}
\bibliography{myref}

@Article{shahriari2016review,
  author    = {Bobak Shahriari and Kevin Swersky and Ziyu Wang and Ryan P. Adams and Nando de Freitas},
  title     = {Taking the Human Out of the Loop: A Review of {Bayesian} Optimization},
  journal   = {Proceedings of the IEEE},
  year      = {2016},
  volume    = {104},
  number    = {1},
  pages     = {148--175},
  doi       = {10.1109/JPROC.2015.2494218},
}

@Article{cruz2024survey,
  author    = {Nayely V\'elez-Cruz},
  title     = {A survey on {Bayesian} nonparametric learning for time series analysis},
  journal   = {Frontiers in Signal Processing},
  year      = {2024},
  volume    = {3},
  pages     = {1287516},
  doi       = {10.3389/frsip.2023.1287516},
}

@Conference{liuxu2021multiresolution,
  author    = {Xu Liu and Decai Li and Yuqing He},
  title     = {Multiresolution Representations for Large-Scale Terrain with Local {Gaussian} Process Regression},
  booktitle = {2021 IEEE International Conference on Robotics and Automation (ICRA 2021)},
  year      = {2021},
  date      = {2021-05-31},
  location  = {Xi'an, China},
  doi       = {10.1109/ICRA48506.2021.9562039},
}

@Article{wangjunyang2021bayesian,
  author    = {Junyang Wang and Jon Cockayne and Oksana Chkrebtii and T. J. Sullivan and Chris. J. Oates},
  title     = {Bayesian numerical methods for nonlinear partial differential equations},
  journal   = {Statistics and Computing},
  year      = {2021},
  volume    = {31},
  pages     = {55},
  doi       = {10.1007/s11222-021-10030-w},
  review    = {The alternative covariance model is inappropriate, causing our PNM to be over-confident.},
}

@Article{ghahramani2015probabilistic,
  author    = {Zoubin Ghahramani},
  title     = {Probabilistic machine learning and artificial intelligence},
  journal   = {Nature},
  year      = {2015},
  date      = {2015-05-28},
  volume    = {521},
  pages     = {452--459},
  doi       = {10.1038/nature14541},
}

@Book{rasmussen2006gp,
  author    = {C.E. Rasmussen and C.K.I. Williams},
  title     = {Gaussian Processes for Machine Learning},
  year      = {2006},
  publisher = {The MIT Press},
  address   = {Cambridge},
}

@Article{raissi2017ml,
  author    = {Maziar Raissi and Paris Perdikaris and George Em Karniadakis},
  title     = {Machine learning of linear differential equations using {Gaussian} processes},
  journal   = {Journal of Computational Physics},
  year      = {2017},
  volume    = {348},
  pages     = {683--693},
  doi       = {10.1016/j.jcp.2017.07.050},
  review    = {yangliu2021bpinn: Vanilla GPR has difficulties in handling the nonlinearities when applied to solve PDEs, leading to restricted applications.},
}

@Article{raissi2018numerical,
  author    = {Maziar Raissi and Paris Perdikaris and George Em Karniadakis},
  title     = {Numerical {Gaussian} processes for time-dependent and nonlinear partial differential equations},
  journal   = {SIAM J. Sci. Comput.},
  year      = {2018},
  volume    = {10},
  number    = {1},
  pages     = {A172--A198},
  doi       = {10.1137/17M1120762},
  review    = {Page A181: Although Gaussian processes can yield satisfactory accuracy, they, by construction, cannot force the approximation error down to machine precision.},
}

@Article{dietrich1997fastgp,
  author    = {C. R. Dietrich and G. N. Newsam},
  title     = {Fast and exact simulation of stationary {Gaussian} processes through circulant embedding of the covariance matrix},
  journal   = {SIAM J. Sci. Comput.},
  year      = {1997},
  volume    = {18},
  number    = {4},
  pages     = {1088--1107},
}

@Article{owhadi2023gph,
  author    = {H. Owhadi},
  title     = {Gaussian process hydrodynamics},
  journal   = {Applied Mathematics and Mechanics},
  year      = {2023},
  volume    = {44},
  number    = {7},
  pages     = {1175--1198},
  doi       = {10.1007/s10483-023-2990-9},
}

@Article{liuhaitao2020review,
  author           = {Liu, Haitao and Ong, Yew-Soon and Shen, Xiaobo and Cai, Jianfei},
  title            = {When {Gaussian} Process Meets Big Data: A Review of Scalable {GPs}},
  journal          = {IEEE Transactions on Neural Networks and Learning Systems},
  year             = {2020},
  volume           = {31},
  number           = {11},
  pages            = {4405--4423},
  issn             = {2162-2388},
  creationdate     = {2024-01-06T22:55:30},
  doi              = {10.1109/tnnls.2019.2957109},
  modificationdate = {2024-01-06T23:47:11},
  publisher        = {Institute of Electrical and Electronics Engineers (IEEE)},
}

@Article{heaton2019competition,
  author    = {Matthew J. Heaton and Abhirup Datta and Andrew O. Finley and Reinhard Furrer and Joseph Guinness and Rajarshi Guhaniyogi and Florian Gerber and Robert B. Gramacy and Dorit Hammerling and Matthias Katzfuss and Finn Lindgren and Douglas W. Nychka and Furong Sun and Andrew Zammit-Mangion},
  title     = {A Case Study Competition Among Methods for Analyzing Large Spatial Data},
  journal   = {Journal of Agricultural, Biological, and Environmental Statistics},
  year      = {2019},
  volume    = {24},
  number    = {3},
  pages     = {398--425},
  doi       = {10.1007/s13253-018-00348-w},
}

@Article{cressie2022basis,
  author    = {Noel Cressie and Matthew Sainsbury-Dale and Andrew Zammit-Mangion},
  title     = {Basis-Function Models in Spatial Statistics},
  journal   = {Annual Review of Statistics and Its Application},
  year      = {2022},
  volume    = {9},
  pages     = {373--400},
  doi       = {10.1146/annurev-statistics-040120-020733},
}

@InProceedings{rasmussen2005healing,
  author    = {Carl Edward Rasmussen and Joaquin Qui{\~n}onero-Candela},
  title     = {Healing the relevance vector machine through augmentation},
  booktitle = {Proceedings of the 22nd International Conference on Machine Learning},
  year      = {2005},
  location  = {Bonn, Germany},
}

@InProceedings{titsias2009sgpr,
  author       = {M. K. Titsias},
  title        = {Variational learning of inducing variables in sparse {Gaussian} processes},
  booktitle    = {Proceedings of the 12th International Conference on Artificial Intelligence and Statistics (AISTATS)},
  year         = {2009},
  volume       = {5},
  organization = {JMLR},
  pages        = {567--574},
  doi          = {10.1007/978-90-481-3723-7-67},
  address      = {Clearwater Beach, Florida, USA},
}

@InProceedings{hensman2013gp,
  author    = {James Hensman and Nicol\'o Fusi and Neil D. Lawrence},
  title     = {Gaussian Processes for Big Data},
  booktitle = {Proc. Conference on Uncertainty in Artificial Intelligence},
  year      = {2013},
}

@Article{csato2002sgp,
  author    = {Lehel Csat\'o and Manfred Opper},
  title     = {Sparse On-Line {Gaussian} Processes},
  journal   = {Neural Computation},
  year      = {2002},
  volume    = {14},
  pages     = {641--668},
  doi       = {10.1162/089976602317250933},
}

@Article{snelson2006spgp,
  author    = {Edward Snelson and Zoubin Ghahramani},
  title     = {Sparse {Gaussian} Processes using Pseudo-inputs},
  journal   = {Advances in Neural Information Processing Systems},
  year      = {2006},
  volume    = {18},
  number    = {1},
  pages     = {1257--1264},
  doi       = {10.1007/s11665-008-9346-x},
}

@InProceedings{ketenci2025cvgp,
  author    = {Mert Ketenci and Adler Perotte and No\'emie Elhadad and I\~nigo Urteaga},
  title     = {Accurate and Scalable Stochastic {Gaussian} Process Regression via Learnable Coreset-based Variational Inference},
  booktitle = {Proceedings of the 41st Conference on Uncertainty in Artificial Intelligence (UAI 2025)},
  year      = {2025},
  volume    = {244},
  pages     = {2101--2142},
}

@Conference{wilson2015kissgp,
  author    = {Andrew Gordon Wilson and Hannes Nickisch},
  title     = {{Kernel interpolation for scalable structured Gaussian processes (KISS-GP)}},
  booktitle = {International Conference of Machine Learning},
  year      = {2015},
}

@Conference{fox2012multiresolution,
  author    = {Emily B. Fox and David B. Dunson},
  title     = {Multiresolution {Gaussian} Processes},
  booktitle = {NIPS'12: Proceedings of the 25th International Conference on Neural Information Processing Systems},
  year      = {2012},
  date      = {2012-12-03},
  location  = {Lake Tahoe, Nevada},
  pages     = {737--745},
}

@Conference{taghia2019multiresolution,
  author    = {Jalil Taghia and Thomas B. Sch\"on},
  title     = {Conditionally Independent Multiresolution {Gaussian} Processes},
  booktitle = {22nd International Conference on Artificial Intelligence and Statistics, AISTATS 2019},
  year      = {2019},
  date      = {2019-04-16},
  publisher = {Scopus},
}

@Article{nychka2015multiresolution,
  author    = {Douglas Nychka and Soutir Bandyopadhyay and Dorit Hammerling and Finn Lindgren and Stephan Sain},
  title     = {A Multiresolution {Gaussian} Process Model for the Analysis of Large Spatial Datasets},
  journal   = {Journal of Computational and Graphical Statistics},
  year      = {2015},
  volume    = {24},
  number    = {2},
  pages     = {579--599},
  doi       = {10.1080/10618600.2014.914946},
}

@Article{gramacy2016lagp,
  author    = {R. B. Gramacy},
  title     = {{LaGP: Large-scale spatial modeling via local approximate Gaussian processes in R}},
  journal   = {J. Stat. Software},
  year      = {2016},
  volume    = {72},
  number    = {1},
  pages     = {1--46},
  doi       = {10.18637/jss.v072.i01},
}

@Article{fuhg2022lagp,
  author    = {Jan N. Fuhg and Michele Marino and Nikolaos Bouklas},
  title     = {Local approximate {Gaussian} process regression for data-driven constitutive models: development and comparison with neural networks},
  journal   = {Comput. Methods Appl. Mech. Engrg.},
  year      = {2022},
  volume    = {388},
  pages     = {114217},
  doi       = {10.1016/j.cma.2021.114217},
}

@Article{katzfuss2021vecchia,
  author    = {Matthias Katzfuss and Joseph Guinness},
  title     = {A General Framework for {Vecchia} Approximations of {Gaussian} Processes},
  journal   = {Statistical Science},
  year      = {2021},
  volume    = {36},
  number    = {1},
  pages     = {124--141},
  doi       = {10.1214/19-STS755},
}

@Article{katzfuss2022svecgp,
  author    = {Matthias Katzfuss and Joseph Guinness and Earl Lawrence},
  title     = {Scaled {Vecchia} approximation for fast computer-model emulation},
  journal   = {SIAM/ASA Journal on Uncertainty Quantification},
  year      = {2022},
  volume    = {10},
  number    = {2},
  pages     = {537--554},
}

@Article{schafer2021compression,
  author    = {Florian Sch\"afer and T. J. Sullivan and Houman Owhadi},
  title     = {Compression, inversion, and approximate {PCA} of dense kernel matrices at near-linear computational complexity},
  journal   = {Multiscale Modeling and Simulation},
  year      = {2021},
  volume    = {19},
  number    = {2},
  pages     = {388--730},
  doi       = {10.1137/19M129526X},
}

@Article{schafer2021sparse,
  author    = {Florian Sch\"afer and Matthias Katzfuss and Houman Owhadi},
  title     = {{Sparse Cholesky factorization by Kullback-Leibler minimization}},
  journal   = {SIAM Journal on Scientific Computing},
  year      = {2021},
  volume    = {43},
  pages     = {A2019--A2046},
  url       = {http://refhub.elsevier.com/S0021-9991(21)00563-5/bib0F2089EA129259D473B3F75E555F2763s1},
}

@Article{chenyifan2025sparse,
  author    = {Yifan Chen and Houman Owhadi and Florian Sch\"afer},
  title     = {{Sparse Cholesky factorization for solving nonlinear PDEs via Gaussian processes}},
  journal   = {Mathematics of Computation},
  year      = {2025},
  volume    = {94},
  number    = {353},
  pages     = {1235--1280},
  doi       = {10.1090/mcom/3992},
}

@Conference{cunningham2023sparse,
  author    = {H. J. Cunningham and D.A. de Souza and S. Takao and M. van der Wilk and M.P. Deisenroth},
  title     = {Actually Sparse Variational {Gaussian} Processes},
  booktitle = {26th International Conference on Artificial Intelligence and Statistics, AISTATS 2023},
  year      = {2023},
  date      = {2023-04-25},
  location  = {Valencia},
}

@Article{katzfuss2017multiresolution,
  author    = {Matthias Katzfuss},
  title     = {A Multi-Resolution Approximation for Massive Spatial Datasets},
  journal   = {Journal of the American Statistical Association},
  year      = {2017},
  volume    = {112},
  number    = {517},
  pages     = {201--214},
  doi       = {10.1080/01621459.2015.1123632},
}

@Article{katzfuss2020multiresolution,
  author    = {Matthias Katzfuss and Wenlong Gong},
  title     = {A class of multi-resolution approximations for large spatial datasets},
  journal   = {Statistica Sinica},
  year      = {2020},
  volume    = {30},
  number    = {4},
  pages     = {2203--2226},
  doi       = {10.5705/ss.202018.0285},
}

@Article{huanghuang2019mrgp,
  author    = {Huang Huang and Lewis R. Blake and Dorit Hammerling},
  title     = {Pushing the Limit: A Hybrid Parallel Implementation of the Multi-resolution Approximation for Massive Data},
  journal   = {arXiv},
  year      = {2019},
  pages     = {1905.00141v1},
}

@Article{tipping2001rvm,
  author    = {M. E. Tipping},
  title     = {Sparse {Bayesian} learning and the relevance vector machine},
  journal   = {Journal of Machine Learning Research},
  year      = {2001},
  volume    = {1},
  pages     = {211--244},
}

@Article{gredilla2010ssgp,
  author    = {M. L\'azaro-Gredilla and J. Qui{\~n}onero-Candela and C. E. Rasmussen and A. R. Figueiras-Vidal},
  title     = {Sparse spectrum {Gaussian} process regression},
  journal   = {Journal of Machine Learning Research},
  year      = {2010},
  volume    = {11},
  pages     = {1865--1881},
  doi       = {10.1109/TITB.2010.2041064},
}

@Article{tan2016ssgp,
  author    = {Linda S. L. Tan and Victor M. H. Ong and David J. Nott and Ajay Jasra},
  title     = {Variational inference for sparse spectrum {Gaussian} process regression},
  journal   = {Statistics and Computating},
  year      = {2016},
  volume    = {26},
  pages     = {1243--1261},
  doi       = {10.1007/s11222-015-9600-7},
}

@Article{cressie2008frk,
  author    = {Noel Cressie and Gardar Johannesson},
  title     = {Fixed rank kriging for very large spatial data sets},
  journal   = {J. R. Statist. Soc. B},
  year      = {2008},
  volume    = {70},
  number    = {1},
  pages     = {209--226},
  doi       = {10.1111/j.1467-9868.2007.00633.x},
}

@Article{mangion2021frk,
  author    = {Andrew Zammit-Mangion and Noel Cressie},
  title     = {{FRK: An R} Package for Spatial and Spatio-Temporal Prediction with Large Datasets},
  journal   = {Journal of Statistical Software},
  year      = {2021},
  volume    = {98},
  number    = {4},
  pages     = {1--48},
  doi       = {10.18637/jss.v098.i04},
}

@Article{greengard2025efgp,
  author    = {Philip Greengard and Manas Rachh and Alex H. Barnett},
  title     = {Equispaced {Fourier} Representations for Efficient {Gaussian} Process Regression from a Billion Data Points},
  journal   = {SIAM/ASA Journal on Uncertainty Quantification},
  year      = {2025},
  volume    = {13},
  number    = {1},
  pages     = {63--89},
  doi       = {10.1137/23M1565310},
}

@Article{kielstra2025gp,
  author    = {P. Michael Kielstra and Michael Lindsey},
  title     = {Gaussian process regression with log-linear scaling for common non-stationary kernels},
  journal   = {Applied and Computational Harmonic Analysis},
  year      = {2025},
  volume    = {79},
  pages     = {101792},
  doi       = {10.1016/j.acha.2025.101792},
}

@Article{datta2016nngp,
  author    = {Abhirup Datta and Sudipto Banerjee and Andrew O. Finley and Alan E. Gelfand},
  title     = {Hierarchical Nearest-Neighbor {Gaussian} Process Models for Large Geostatistical Datasets},
  journal   = {Journal of the American Statistical Association},
  year      = {2016},
  volume    = {111},
  number    = {514},
  pages     = {800--812},
  doi       = {10.1080/01621459.2015.1044091},
}

@Article{finley2019nngp,
  author    = {Andrew O. Finley and Abhirup Datta and Bruce D. Cook and Douglas C. Morton and Hans E. Andersen and Sudipto Banerjee},
  title     = {Efficient Algorithms for {Bayesian} Nearest Neighbor {Gaussian} Processes},
  journal   = {Journal of Computational and Graphical Statistics},
  year      = {2019},
  volume    = {28},
  number    = {2},
  pages     = {401--414},
  doi       = {10.1080/10618600.2018.1537924},
}

@Article{finley2022spnngp,
  author    = {Andrew O. Finley and Abhirup Datta and Sudipto Banerjee},
  title     = {{spNNGP R} Package for Nearest Neighbor {Gaussian} Process Models},
  journal   = {Journal of Statistical Software},
  year      = {2022},
  volume    = {103},
  pages     = {5},
  doi       = {10.18637/jss.v103.i05},
}

@Article{shaby2012tapering,
  author    = {Benjamin Shaby and David Ruppert},
  title     = {Tapered Covariance: {Bayesian} Estimation and Asymptotics},
  journal   = {Journal of Computational and Graphical Statistics},
  year      = {2012},
  volume    = {21},
  number    = {2},
  pages     = {433--452},
  doi       = {10.1080/10618600.2012.680819},
}

@Article{furrer2006tapering,
  author    = {Reinhard Furrer and Marc G Genton and Douglas Nychka},
  title     = {Covariance Tapering for Interpolation of Large Spatial Datasets},
  journal   = {Journal of Computational and Graphical Statistics},
  year      = {2006},
  volume    = {15},
  number    = {3},
  pages     = {502--523},
  doi       = {10.1198/106186006X132178},
}

@Article{ambikasaran2016gp,
  author    = {Sivaram Ambikasaran and Daniel Foreman-Mackey and Leslie Greengard and David W. Hogg and Michael O'Neil},
  title     = {Fast Direct Methods for {Gaussian} Processes},
  journal   = {IEEE Transactions on Pattern Analysis and Machine Intelligence},
  year      = {2016},
  volume    = {38},
  number    = {2},
  pages     = {252--265},
  doi       = {10.1109/TPAMI.2015.2448083},
}

@Article{guinness2018permutation,
  author    = {Joseph Guinness},
  title     = {Permutation and Grouping Methods for Sharpening {Gaussian} Process Approximations},
  journal   = {Technometrics},
  year      = {2018},
  volume    = {60},
  number    = {4},
  pages     = {415--429},
  doi       = {10.1080/00401706.2018.1437476},
}

@Article{brandhorst2011sparse,
  author    = {Kai Brandhorst and Martin Head-Gordon},
  title     = {Fast Sparse {Cholesky} Decomposition and Inversion using Nested Dissection Matrix Reordering},
  journal   = {Journal of Chemical Theory and Computation},
  year      = {2011},
  volume    = {7},
  pages     = {351--368},
  doi       = {10.1021/ct100618s},
}

@Article{liuwh1986cholesky,
  author    = {Joseph W. H. Liu},
  title     = {A compact row storage scheme for {Cholesky} factors using elimination trees},
  journal   = {ACM Transactions on Mathematical Software},
  year      = {1986},
  volume    = {12},
  number    = {2},
  pages     = {127--148},
}

@Article{noack2023nonstationary,
  author    = {Marcus M. Noack and Harinarayan Krishnan and Mark D. Risser and Kristofer G. Reyes},
  title     = {Exact {Gaussian} processes for massive datasets via non-stationary sparsity-discovering kernels},
  journal   = {Scientific Reports},
  year      = {2023},
  volume    = {13},
  pages     = {3155},
  doi       = {10.1038/s41598-023-30062-8},
}

@Conference{dongkun2017determinants,
  author    = {Kun Dong and David Eriksson and Hannes Nickisch and David Bindel and Andrew Gordon Wilson},
  title     = {Scalable Log Determinants for {Gaussian} Process Kernel Learning},
  booktitle = {31st Conference on Neural Information Processing Systems (NIPS 2017)},
  year      = {2017},
  address   = {Long Beach, CA, USA},
}

@Article{gardner2018skip,
  author    = {Jacob R. Gardner and Geoff Pleiss and Ruihan Wu and Kilian Q. Weinberger and Andrew Gordon Wilson},
  title     = {Product Kernel Interpolation for Scalable {Gaussian} Processes},
  journal   = {arXiv},
  year      = {2018},
  pages     = {1802.08903},
}

@Conference{banhanyuan2024mkissgp,
  author    = {Hanyuan Ban and Ellen H. J. Riemens and Raj Thilak Rajan},
  title     = {Malleable Kernel Interpolation for Scalable Structured {Gaussian} Process},
  booktitle = {32nd European Signal Processing Conference (EUSIPCO)},
  year      = {2024},
  doi       = {10.23919/EUSIPCO63174.2024.10715101},
  isbn      = {978-9-4645-9361-7},
}

@Article{minden2017sgp,
  author    = {Victor Minden and Anil Damle and Kenneth L. Ho and Lexing Ying},
  title     = {Fast spatial {Gaussian} process maximum likelihood estimationvia skeletonization factorizations},
  journal   = {Multiscale Modeling and Simulation},
  year      = {2017},
  volume    = {15},
  number    = {4},
  pages     = {1584--1311},
  url       = {http://refhub.elsevier.com/S1063-5203(25)00046-6/bib57A27EFFE958EEE9FAE2158DB1199C0Ds1},
}

@Article{tuong2009lgp,
  author    = {Duy Nguyen-Tuong and Matthias Seeger and Jan Peters},
  title     = {Model Learning with Local {Gaussian} Process Regression},
  journal   = {Advanced Robotics},
  year      = {2009},
  volume    = {23},
  pages     = {2015--2034},
  doi       = {10.1163/016918609X12529286896877},
}

@Conference{gaoyinghua2020lgp,
  author    = {Yinghua Gao and Naiqi Li and Ning Ding and Yiming Li and Tao Dai and Shu-Tao Xia},
  title     = {Generalized Local Aggregation for Large Scale {Gaussian} Process Regression},
  booktitle = {International Joint Conference on Neural Networks (IJCNN)},
  year      = {2020},
  address   = {Glasgow, United Kingdom, United Kingdom},
  date      = {2020-07-19},
  doi       = {10.1109/IJCNN48605.2020.9207107},
}

@Article{gramacy2015lagp,
  author    = {Robert B. Gramacy and Daniel W. Apley},
  title     = {Local {Gaussian} Process Approximation for Large Computer Experiments},
  journal   = {Journal of Computational and Graphical Statistics},
  year      = {2015},
  volume    = {24},
  number    = {2},
  pages     = {561--578},
  doi       = {10.1080/10618600.2014.914442},
}

@Article{rumsey2023leapgp,
  author    = {Kellin N. Rumsey and Gabriel Huerta and J. Derek Tucker},
  title     = {A localized ensemble of approximate {Gaussian} processes for fast sequential emulation},
  journal   = {Stat.},
  year      = {2023},
  volume    = {12},
  pages     = {e576},
  doi       = {10.1002/sta4.576},
}

@Article{adjetey2026jump,
  author    = {Isaac Adjetey and Yiyuan She},
  title     = {Mind the jumps: A scalable robust local {Gaussian} process for multidimensional response surfaces with discontinuities},
  journal   = {Neurocomputing},
  year      = {2026},
  volume    = {667},
  pages     = {132317},
  doi       = {10.1016/j.neucom.2025.132317},
}

@Article{wangke2019exactgp,
  author    = {Ke Alexander Wang and Geoff Pleiss and Jacob R. Gardner and Stephen Tyree and Kilian Q. Weinberger and Andrew Gordon Wilson},
  title     = {Exact {Gaussian} Processes on a Million Data Points},
  journal   = {Adv. Neural. Inf. Process. Syst.},
  year      = {2019},
  volume    = {32},
  pages     = {14648--14659},
}

@Article{rumsey2025emulators,
  author    = {Kellin N. Rumsey and Graham C. Gibson and Devin Francom and Reid Morris},
  title     = {All Emulators are Wrong, Many are Useful, and Some are More Useful Than Others: A Reproducible Comparison of Computer Model Surrogates},
  journal   = {arXiv},
  year      = {2025},
  pages     = {2512.09060},
}

@Article{manfredi2024uncertainty,
  author    = {P. Manfredi},
  title     = {Probabilistic uncertainty propagation using {Gaussian} Process surrogates},
  journal   = {International Journal for Uncertainty Quantification},
  year      = {2024},
  volume    = {14},
  number    = {6},
  pages     = {71--104},
}

@Article{lijinglai2025gpr4uq,
  author    = {Jinglai Li and Hongqiao Wang},
  title     = {Gaussian Processes Regression for Uncertainty Quantification: An Introductory Tutorial},
  journal   = {arXiv},
  year      = {2025},
  pages     = {2502.03090v2},
}

@Article{wenger2022uncertainty,
  author    = {Jonathan Wenger and Geoff Pleiss and Marvin Pf\"ortner and Philipp Hennig and John P. Cunningham},
  title     = {Posterior and Computational Uncertainty in {Gaussian} Processes},
  journal   = {arXiv},
  year      = {2022},
  pages     = {2205.15449},
}

@InProceedings{wuluhuan2022vnngp,
  author    = {Luhuan Wu and Geoff Pleiss and John Cunningham},
  title     = {Variational Nearest Neighbor {Gaussian} Processes},
  booktitle = {Proceedings of the 39th International Conference on Machine Learning, PMLR 162},
  year      = {2022},
  address   = {Baltimore, Maryland, USA},
  date      = {2022-07-17},
}

@Article{guohongwei2026pigp,
  author    = {Hongwei Guo and Zhen-Yu Yin},
  title     = {{Physics-informed continuous and discrete Gaussian processes for forward and inverse soil consolidation analysis: Uncertainty quantification and diffusion kernel-based multi-fidelity modeling}},
  journal   = {Computer Methods in Applied Mechanics and Engineering},
  year      = {2026},
  volume    = {448},
  pages     = {118438},
  doi       = {10.1016/j.cma.2025.118438},
}

\nolinenumbers

\end{document}